\pdfoutput=1

\documentclass[11pt]{article}

\usepackage[final]{acl}

\usepackage{amsmath,amsfonts,bm}

\def\eqref#1{equation~\ref{#1}}

\def\1{\bm{1}}

\DeclareMathAlphabet{\mathsfit}{\encodingdefault}{\sfdefault}{m}{sl}
\SetMathAlphabet{\mathsfit}{bold}{\encodingdefault}{\sfdefault}{bx}{n}

\usepackage{graphicx}
\usepackage{multirow}
\usepackage{algorithm}
\usepackage{algorithmic}
\usepackage{amsfonts}
\usepackage{times}
\usepackage{latexsym}

\usepackage[T1]{fontenc}
\usepackage{graphicx}

\usepackage[utf8]{inputenc}

\usepackage{microtype}

\usepackage{inconsolata}

\title{Speech language models lack important brain-relevant semantics}

\author{Subba Reddy Oota$^{1,2}$, Emin Çelik$^{2}$, Fatma Deniz$^{3}$, Mariya Toneva$^2$\\
$^1$Inria Bordeaux, France, \\
$^2$Max Planck Institute for Software Systems, Germany, \\ 
$^3$Technische Universität Berlin, Germany\\
  \scriptsize \texttt{subba-reddy.oota@inria.fr, emcelik@mpi-sws.org, deniz@tu-berlin.de, mtoneva@mpi-sws.org} }

\begin{document}
\maketitle
\begin{abstract}
Despite known differences between reading and listening in the brain, recent work has shown that text-based language models predict both text-evoked and speech-evoked brain activity to an impressive degree. This poses the question of what types of information language models truly predict in the brain. We investigate this question via a direct approach, in which we systematically remove specific low-level stimulus features (textual, speech, and visual) from language model representations to assess their impact on alignment with fMRI brain recordings during reading and listening. Comparing these findings with speech-based language models reveals starkly different effects of low-level features on brain alignment. While text-based models show reduced alignment in early sensory regions post-removal, they retain significant predictive power in late language regions. In contrast, speech-based models maintain strong alignment in early auditory regions even after feature removal but lose all predictive power in late language regions. These results suggest that speech-based models provide insights into additional information processed by early auditory regions, but caution is needed when using them to model processing in late language regions. We make our code publicly available.~\footnote{\url{https://github.com/subbareddy248/speech-llm-brain}}
\end{abstract}

\section{Introduction}

An explosion of recent work that investigates the alignment between the human brain and language models shows that text-based language models (e.g. GPT*, BERT, etc.) predict both text- and speech-evoked brain activity to an impressive degree (text: \citep{toneva2019interpreting,schrimpf2021neural,goldstein2022shared,aw2022training,oota2022neural,lamarre2022attention,chen2024cortical}; speech: \citep{jain2018incorporating,caucheteux2020language,antonello2021low,vaidya2022self,millet2022toward,tuckute2022many,oota2022joint,oota2023neural,oota2023speech,chen2024cortical}). This observation holds across late language regions, which are thought to process both text- and speech-evoked language~\citep{deniz2019representation}, but also more surprisingly across early sensory cortices, which are shown to be modality-specific~\citep{deniz2019representation,chen2024cortical}. Since text-based language models are trained on written text~\citep{devlin2019bert,radford2019language,chung2022scaling}, their impressive performance at predicting the activity in (also referred to as alignment with) early auditory cortices is puzzling. This raises the question of what types of information underlie the brain alignment of language models observed across brain regions.

\begin{figure*}[t] 
\centering
\begin{minipage}{0.44\textwidth}
\centering
\includegraphics[width=\linewidth]    {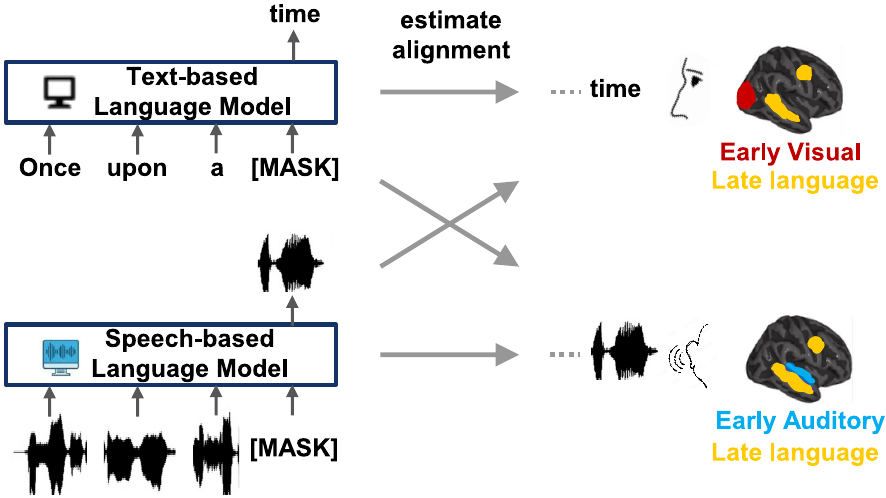}
\end{minipage}
\vline
\begin{minipage}{0.55\textwidth}
\centering
\includegraphics[width=\linewidth]{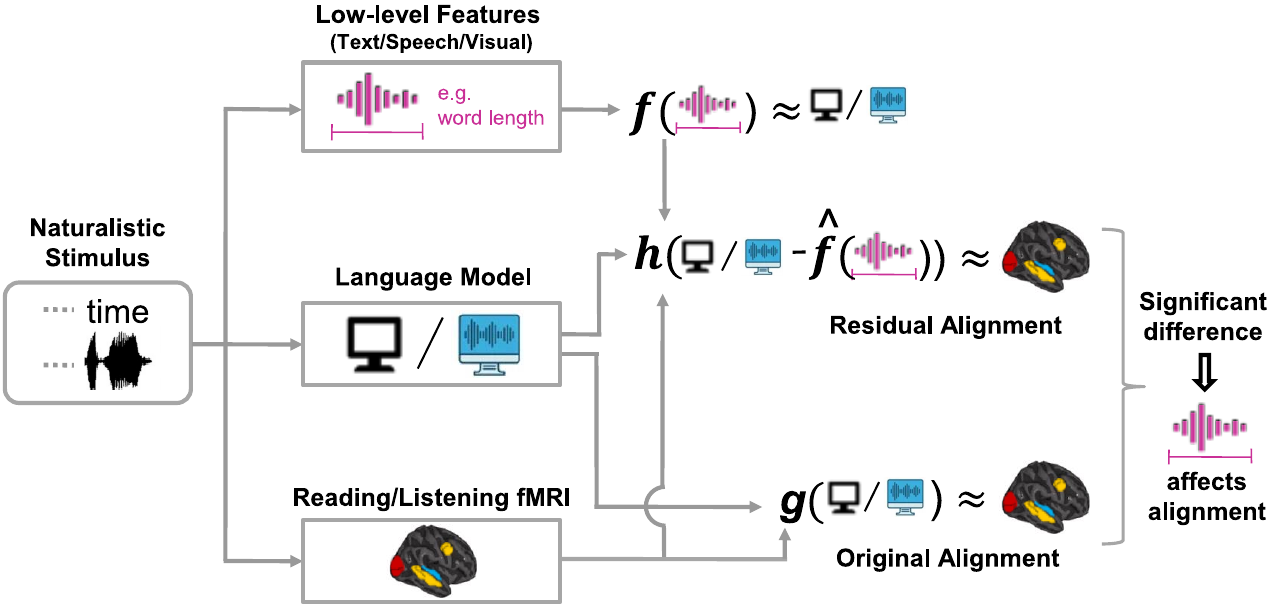}
\end{minipage}
\caption{A direct approach to test the effect of low-level stimulus features on the alignment between different types
of language models and brain recordings (reading vs. listening).}
\label{fig:approach}
\end{figure*}

In this work, we investigate this question via a direct approach (see Fig.~\ref{fig:approach} for a schematic). For a number of low-level textual, speech, and visual features, we analyze how the alignment between brain recordings and language model representations is affected by the elimination of information related to these features. We refer to this approach as direct, because it estimates the direct effect of a specific feature on brain alignment. This approach is in contrast to an indirect approach, which first estimates the brain alignment of a model and then independently examines the model's performance on a natural language processing task (e.g. approaches presented by \citet{schrimpf2021neural,goldstein2022shared}). In contrast, our approach can directly estimate the impact of a specific feature on the alignment between the model and the brain recordings by observing the difference in alignment before and after the specific feature is computationally removed from the model representations.

We further contrast our findings with speech-based language models, which would be expected to predict speech-evoked brain activity better, provided they model language processing in the brain well. 
For this purpose, we present a systematic study of the brain alignment across two popular fMRI datasets of naturalistic stories (1-reading, 1-listening) and different natural language processing models (text vs. speech).
We focus on three popular text-based language models (BERT~\citep{devlin2019bert},  GPT2~\citep{radford2019language}, and FLAN-T5~\citep{chung2022scaling}) and two speech-based models (Wav2vec2.0~\citep{baevski2020wav2vec} and Whisper~\citep{radford2022robust})–which have been studied extensively in the NLP-brain alignment literature~\citep{toneva2019interpreting,aw2022training,merlin2022language,oota2022neural,oota2022joint,millet2022toward,vaidya2022self}.
The fMRI recordings are openly
available~\citep{deniz2019representation} and correspond to 6 participants reading and listening to the same naturalistic stories. This dataset is unique in presenting the same naturalistic stimuli under two sensory modalities. Ensuring the same stimuli across sensory modalities is crucial, as differences in stimuli could otherwise account for any observed differences in brain recordings across modalities. We test how eliminating a comprehensive set of low-level textual (number of letters, number of words, word length, etc.), speech (number of phonemes, Fbank, MFCC, Mel, articulation, phonological features, etc.), and visual features (motion energy) from model representations affects alignment with brain responses.

Our direct approach leads to three important conclusions: (1) The surprising alignment of models with sensory regions corresponding to the incongruent modality (i.e. text models with auditory regions and speech models with visual regions) is entirely due to low-level stimulus features that are correlated between text and speech (e.g. number of letters and number of phonemes); (2) Models align differently with their corresponding sensory regions, and this difference is not explained by low-level stimulus features. Text models exhibit comparable alignment with visual and auditory regions, entirely explained by low-level textual features. In contrast, speech models display significantly greater alignment with auditory regions than visual regions, and this difference cannot be entirely accounted for by the comprehensive set of low-level stimulus features studied. This underscores the capacity of speech-based language models to capture additional features crucial for early auditory cortex, suggesting their potential in enhancing our understanding of this brain region; (3) Although both text- and speech-based models exhibit considerable alignment with late language regions, the influence of low-level stimulus features on text model alignment is minimal, whereas for speech-based models, the alignment is entirely explained by these low-level features. This disparity raises questions about the utility of speech-based models in modeling late language processing, contrasting with the reassuring findings for text models. 
Our work demonstrates the importance of carefully considering the reasons behind
the observed alignment between models and human brain recordings.
We make all code available so that others can reproduce and build on our methodology and findings.

\section{Related Work}

\label{sec:related-work}
Our work is most closely related to that of~\citet{toneva2022combining}, who propose the direct residual approach to study the supra-word meaning of language by removing the contribution of individual words to brain alignment.
More recent work uses the same residual approach to investigate the effect of syntactic and semantic properties on brain alignment across layers of a text-based language model~\citep{oota2022joint}. We complement these works by studying the impact of a wide range of low-level features on brain alignment, and by examining this effect on alignment with both text and speech models. 

Other direct approaches have also been proposed in the literature. Most notably, work by \citet{ramakrishnan2021non} studies the impact of removing information related to word embeddings directly from brain responses on a downstream task. Conceptually, the results obtained from this approach and ours should be similar because the feature is completely removed from either the brain alignment input, target, or both, and thus cannot further impact the observed alignment. However, in practice, brain recordings are noisier than model representations, making it more challenging to estimate the removal regression model, especially with a limited sample size. Therefore, in our work, we opt to remove features from the model representations rather than from the brain recordings. Methods based on variance partitioning~\citep{lescroart2015fourier,deniz2019representation} offer another direct approach. Our approach is complementary and easily allows relating the feature effect to an estimated noise ceiling, given by cross-subject prediction.

Our work also relates to a growing literature that investigates the alignment between human brains and language models. 
A number of studies have used text-based language models to predict both text-evoked and speech-evoked brain activity to an impressive degree~\citep{wehbe2014simultaneously,jain2018incorporating,toneva2019interpreting,schwartz2019inducing,caucheteux2020language,jat2020relating,abdou2021does,toneva2022same,antonello2021low,oota2022neural,merlin2022language,aw2022training,oota2022joint,lamarre2022attention}.
Similarly, the recent advancements in Transformer-based models for speech~\citep{chung2020vector,baevski2020wav2vec,hsu2021hubert} have motivated neuroscience researchers to test their brain alignment with speech-evoked brain activity~\citep{millet2022toward,vaidya2022self,tuckute2022many,oota2023neural,oota2023speech,chen2024cortical}.
Our approach is complementary and can be used to further understand what types of information underlie the brain alignment of language models, particularly across different brain regions. 

\section{Datasets and Models}

\subsection{Brain Datasets}

We analyzed two fMRI datasets which were recorded while the same 6 participants listened to and read the same narrative stories from the Moth Radio Hour, with 3737 training and 291 testing samples (TRs: Time Repetition) each, chosen for the unique auditory and visual presentation of the same stimulus. 
These publicly available datasets, provided by~\citet{deniz2019representation}, were examined using the Glasser Atlas' multi-modal parcellation of the cerebral cortex, targeting 180 ROIs per hemisphere~\citep{glasser2016multi}. 
This includes two early sensory processing regions (early visual and early auditory) and seven language-relevant regions, encompassing broader language regions: angular gyrus, lateral temporal cortex, inferior frontal gyrus and middle frontal gyrus, based on functional localizers of the language network~\citep{fedorenko2010new,fedorenko2014reworking,milton2021parcellation,desai2022proper}.
More details about the dataset and ROI selectivity are reported in the Appendix (see Section~\ref{naturalistic_dataset} and Table~\ref{rois_description}).

\begin{figure*}[t] 
\centering
\includegraphics[width=\linewidth]{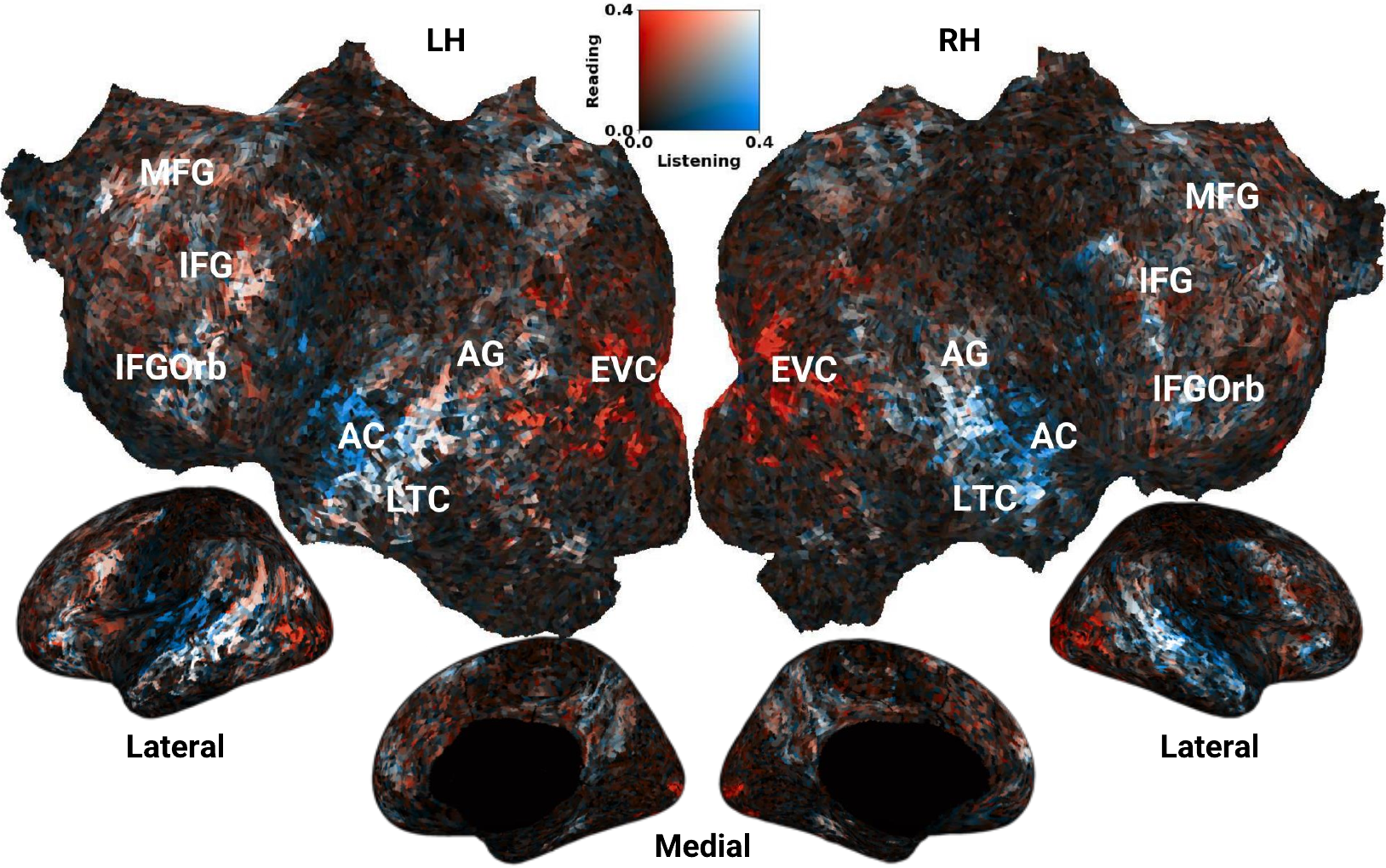}
\caption{Contrast of estimated cross-subject prediction accuracy for reading and listening for a representative subject (subject-8). 
\textcolor{cyan}{Blue} and \textcolor{red}{Red} voxels depict higher cross-subject prediction accuracy estimates during listening and reading, respectively. Voxels that have similar cross-subject prediction accuracy during reading and listening appear white, and are distributed across language regions. Here, middle frontal gyrus (MFG), inferior frontal gyrus (IFG),  inferior frontal gyrus orbital (IFGOrb), angular gyrus (AG), and lateral temporal cortex (LTC) are late language regions, EVC denotes early visual cortex and AC denotes auditory cortex.
Cross-subject prediction accuracy for other participants are reported in Appendix Figs.~\ref{fig:noise_ceiling_subject01} and ~\ref{fig:noise_ceiling_subject07}. }
\label{fig:noise_ceiling_subject05}
\end{figure*}

\noindent\textbf{Estimating dataset cross-subject prediction accuracy.}
To account for the intrinsic noise in biological measurements, we adapt a previously proposed method to estimate the noise ceiling value (i.e. cross-subject prediction accuracy) for a model's performance for the reading and listening fMRI datasets \citep{schrimpf2021neural}. 
By subsampling fMRI datasets from 6 participants, we generate all possible combinations of $s$ participants ($s$ $\in$ [2,6]) for both reading and listening tasks, and use a voxel-wise encoding model (see Sec. \ref{sec:methods}) to predict one participant's response using others.
Note that the estimated cross-subject prediction accuracy is based on the assumption of a perfect model, which might differ from real-world scenarios, yet offers valuable insights into a model's performance.
We present the average cross-subject prediction accuracy across voxels for the \emph{reading-listening fMRI} dataset in Appendix Fig.~\ref{fig:noise_ceiling_datasets}, which shows no significant differences between modalities.
However, clear differences in prediction accuracy can be observed on the region level in Fig.~\ref{fig:noise_ceiling_subject05}: early visual regions are better predicted during reading (\textcolor{red}{Red}) and early auditory regions during listening (\textcolor{cyan}{Blue}).

\vspace{-0.1cm}
\subsection{Language Models}
\vspace{-0.1cm}

To investigate the reasons for brain alignment of language models during reading and listening, we extract activations from five popular pretrained Transformer models. Three of these models are ``text-based" and two are ``speech-based". Below we present more details for each model.

\begin{table}[t]
\scriptsize
\begin{center} 

\resizebox{0.5\textwidth}{!}{\begin{tabular}{|c|c|c|c|} 
\hline
\textbf{Model Name} & \textbf{Pretraining } & \textbf{Type} & \textbf{Layers} \\
\hline
BERT-base-uncased  & Text&  Encoder (Bidirectional) & 12  \\
GPT2-small &Text &   Decoder (Unidirectional) & 12 \\
FLAN-T5-base& Text&Encoder-Decoder & 24\\
Wav2Vec2.0-base&Speech& Encoder &  12\\
Whisper-small&Speech-to-Text& Encoder-Decoder &  24\\
\hline
\end{tabular} }
\caption{Pretrained Transformer-based language models} 
\label{neural_models} 
\end{center} 
\end{table}

\noindent\textbf{Text-based language models.}
To extract representations of the text stimulus, we use three popular pretrained Transformer text-based language models from Huggingface~\citep{wolf2020transformers}: (1) BERT~\citep{devlin2019bert}, 
(2) GPT-2~\citep{radford2019language} and 
(3) FLAN-T5~\citep{chung2022scaling}. We report details of each model in Table~\ref{neural_models}. 


\noindent\textbf{Extracting text representations.}
We follow previous work to extract the hidden-state representations from each layer of these language models, given a fixed-length input length~\citep{toneva2019interpreting}. To extract the stimulus features from these pretrained models, we constrained the tokenizer to use a maximum context of 20 words. 
Given the constrained context length, each word is successively input to the network with at most $C$ previous tokens.
For instance, given a story of $M$ words and considering the context length of 20, while the third word’s vector is computed by presenting (w$_1$, w$_2$, w$_3$) as input to the network, the last word’s vector w$_M$ is computed by presenting the network with (w$_{M-20}$, $\dots$, w$_M$). 
The pretrained Transformer model outputs token representations at different layers. We use the \#tokens $\times$ 768 dimension vector obtained from each hidden layer to obtain word-level representations from each pretrained Transformer language model. 
To align the stimulus presentation rate with the slower fMRI data acquisition rate (TR = 2.0045 sec), where multiple words correspond to a single TR, we downsample the stimulus features to match fMRI recording times using a 3-lobed Lanczos filter, thus creating chunk-embeddings for each TR.


\noindent\textbf{Speech-based language models.}
Similarly to text-based language models, we use two popular pretrained Transformer speech-based models from Huggingface: (1) Wav2Vec2.0~\citep{baevski2020wav2vec} and (2) Whisper ~\citep{radford2022robust}. 
The details of each model are reported in Table~\ref{neural_models} and Appendix Table~\ref{neural_models_details}.




\noindent\textbf{Extracting speech representations.}
The input audio story is first segmented into clips corresponding to 2.0045 seconds to match the fMRI image acquisition rate. Each audio clip is input to the speech-based models one by one to obtain stimulus representations per clip. The representations are obtained from the activations of the pretrained speech model in intermediate layers. 
Overall, each layer of the examined speech-based models outputs a 768 dimensional vector at each TR. 

To investigate whether speech models incorporate linguistic information beyond 2 seconds, we extract representations for longer timewindows (16, 32, and 64 secs) with a stride of 100 msecs and considered the last token as representation in each context window.  
Similarly to text-based language models, the pretrained speech-based models output token representations at different layers. 
Finally, we align these representations with the fMRI data by downsampling the stimulus features with a 3-lobed Lanczos filter, thus producing chunk-embeddings for each TR.

\vspace{-0.1cm}
\subsection{Interpretable Stimulus Features}
\vspace{-0.1cm}

To better understand the contribution of different stimulus features to the brain alignment of language models, we extract a range of low-level textual, speech, and visual features that have been shown in previous work to relate to brain activity during listening and reading. 


\noindent\textbf{Low-level textual features.} 
We analyze low-level textual features per TR, including the (1) \emph{Number of Letters}, (2) \emph{Number of Words}, and (3) \emph{Word Length STD}.
These low-level textual features, which are already downsampled and aligned with each TR, have also been used in ~\citet{deniz2019representation}.

\noindent\textbf{Low-level speech features.}
We consider the following low-level speech features:
(1) \emph{Number of Phonemes} in each TR. (2) \emph{MonoPhones}, the smallest speech units like /p/, /c/, /a/, differentiate words, represented by a 39-dimensional feature vector at each TR.
(3) \emph{DiPhones} represent the adjacent pair of phones (e.g., [da], [aI], [If]) in an utterance. For each TR, we obtained a one-hot code encoding the presence or absence of all possible diphones (858).
(4) \emph{FBank} splits the raw audio into sub-components via bandpass filtering, yielding a 26-dimensional vector for each frequency sub-band.
(5) \emph{Mel Spectrogram} transforms audio signals into an 80-dimensional vector by applying Fourier transformation and mapping frequencies to the mel scale.
(7) \emph{MFCC} features capture Mel-frequency spectral coefficients through Discrete Cosine Transform (DCT) of logarithmic filter bank outputs.
(8) \emph{PowSpec}, detailed in~\citet{gong2023phonemic}, quantify the time-varying power spectrum of audio signals across 448 frequency bands between 25 Hz and 15 kHz for every 2 sec segment.
(9) \emph{Phonological} features identify 108 phonological aspects (18 descriptors like vocalic, consonantal, back) across 6 statistical functions (mean, std, skewness, kurtosis, max, min).
(10) \emph{Articulation}: we use phoneme articulations as mid-level speech features, mapping hand-labeled phonemes to 22 articulatory characteristics.
We extract the low-level speech features filter banks (FBank), Mel Spectrogram, and MFCC from audio files using S3PRL toolkit\footnote{\url{https://github.com/s3prl/s3prl}}, and phonological features using the DisVoice library\footnote{\url{https://github.com/jcvasquezc/DisVoice}}. We also use the articulation and power spectrum (PowSpec) feature vectors provided in~\citet{deniz2019representation}.

\noindent\textbf{Low-level visual features.}
As low-level visual features, we consider the \emph{Motion energy} features derived from word frame sequences in reading experiments using a spatiotemporal Gabor pyramid. These features capture low-level visual characteristics with 39 parameters, as detailed in~\citet{deniz2019representation}.


\vspace{-0.1cm}
\section{Methodology}
\label{sec:methods}
\vspace{-0.1cm}

Our direct approach to investigate the reasons for brain alignment of language models involves three main steps (see Fig.~\ref{fig:approach}): (1) removal of interpretable low-level stimulus features from the language model representations; (2) estimating the brain alignment of the language model representations before and after removal of a particular feature; (3) a significance test to conclude whether the difference in estimated brain alignment before and after is significant.

\noindent\textbf{Removal of low-level features from language model representations.}
To remove low-level features from language model representations, we rely on a simple method proposed previously by \citet{toneva2022combining}, in which the linear contribution of the feature to the language model activations is removed via ridge regression. In our setting, we remove the linear contribution of a low-level feature by training a ridge regression, in which the low-level feature vector is considered as input and the neural word/speech representations are the target. We compute the residuals by subtracting the predicted feature representations from the actual features resulting in the (linear) removal of low-level feature vector from pretrained features. Because the brain prediction method is also a linear function (see next paragraph), this linear removal limits the contribution of low-level features to the eventual brain alignment.

Specifically, given an input feature vector $\mathbf{L}_{i}$ with dimension  $\mathbf{N}\times \mathbf{d}$ for low-level feature $i$, and target neural model representations $\mathbf{W} \in \mathbb{R}^{\mathbf{N}\times \mathbf{D}}$, where $\mathbf{N}$ denotes the number of TRs, and d and D denote the dimensionality of low-level and neural model representations, respectively, the ridge regression objective function is
     $f(\mathbf{L}_{i}) = \underset{\theta_{i}}{\text{min}} \lVert \mathbf{W} - \mathbf{L}_{i}\theta_{i} \rVert_{F}^{2} + \lambda \lVert \theta_{i} \rVert_{F}^{2}$ 
where $\theta_{i}$ denotes the learned weight coefficient for embedding dimension $\mathbf{D}$ for the input feature $i$, $\lVert.\rVert_{F}^2$ denotes the Frobenius norm, and $\lambda >0$ is a tunable hyper-parameter representing the regularization weight for each feature dimension.
Using the learned weight coefficients, we compute the residuals as follows: $r(\mathbf{L}_{i}) =  \mathbf{W} - \mathbf{L}_{i}\theta_{i}$. 


\noindent\textbf{TR alignment.}
To account for the slowness of the hemodynamic response, we model hemodynamic response function using a finite response filter (FIR) per voxel and for each subject separately with 6 temporal delays corresponding to 12 secs. 

\noindent\textbf{Voxel-wise encoding model.}
We estimate the brain alignment of a language model before and after the removal of a stimulus property via training standard voxel-wise encoding models~\cite{deniz2019representation,toneva2019interpreting}. Specifically, for each voxel and participant, we train fMRI encoding model using bootstrap ridge regression~\citep{tikhonov1977solutions} to predict the fMRI recording associated with this voxel as a function of the stimulus representations obtained from both text- and speech-based models (before and after the removal of stimulus features). 
In particular, we use layerwise pretrained representations from these models as well as residuals by removing each basic low-level feature and using them in a voxelwise encoding model to predict brain responses.
If the removal of a particular stimulus property from the language model representation leads to a significant drop in brain alignment, then we conclude that this stimulus property is important for the brain alignment of the language model.
Before the bootsrap ridge regression, we first z-scored each feature channel separately for training and testing. This was done to match the features to the fMRI responses, which were also z-scored for training and testing.
Formally, at the time step (t), we encode the stimuli as $X_{t}\in \mathbb{R}^{N \times D}$ and brain region voxels $Y_{t}\in \mathbb{R}^{N \times V}$, where $N$ is the number of training examples, $D$ denotes the dimension of the concatenation of delayed 6 TRs, and $V$ denotes the number of voxels.
To find the optimal regularization parameter for each feature space, we use a range of regularization parameters that is explored using cross-validation. The main goal of each fMRI encoding model is to predict brain responses associated with each brain voxel given a stimulus.

\noindent\textbf{Normalized alignment.}
The final measure of a model's performance is obtained by calculating Pearson's correlation between the model's predictions and neural recordings. 
This correlation is then divided by the estimated cross-subject prediction accuracy and averaged across voxels, regions, and participants, resulting in a standardized measure of performance referred to as normalized alignment.
During normalized alignment, we select the voxels whose cross-subject prediction accuracy is $\ge$ 0.05.

\noindent\textbf{Statistical Significance.}
To determine if normalized predictivity scores are significantly higher than chance, we use block permutation tests. We employ the standard implementation of a block permutation test for fMRI data, which is to split the fMRI data into blocks of 10 contiguous TRs and permute the order of these blocks, while maintaining the original order of the TRs within each block. 
By permuting predictions 5000 times, we create an empirical distribution for chance performance, from which we estimate the p-value of the actual performance.
To estimate the statistical significance of performance differences, such as between the model's predictions and chance or residual predictions and chance, we utilized the Wilcoxon signed-rank test, applying it to the mean normalized predictivity for the participants.
In all cases, we denote significant differences with an asterisk \textcolor{red}{*}, indicating cases where p$\leq 0.05$.


\noindent\textbf{Implementation details for reproducibility.}
All experiments were conducted on a machine with 1 NVIDIA GEFORCE-GTX GPU with 16GB GPU RAM. We used bootstrap ridge-regression with the following parameters: MSE loss function, and L2-decay ($\lambda$) varied from  10$^{1}$ to 10$^{3}$; the best $\lambda$ was chosen by tuning on validation data. 

\begin{figure*}[t] 
\centering
\begin{minipage}{0.49\textwidth}
\centering
\includegraphics[width=\linewidth]{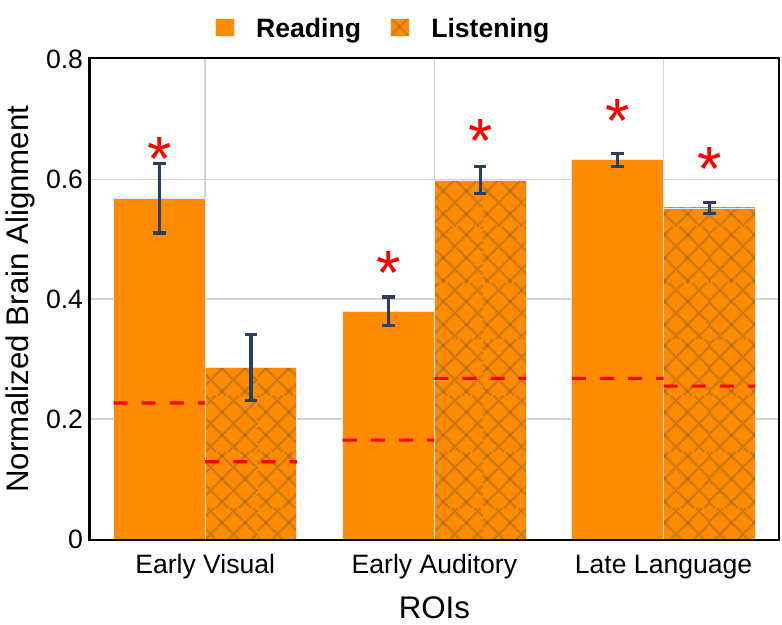}
\\(a) Text Models
\end{minipage}
\begin{minipage}{0.49\textwidth}
\centering
\includegraphics[width=\linewidth]{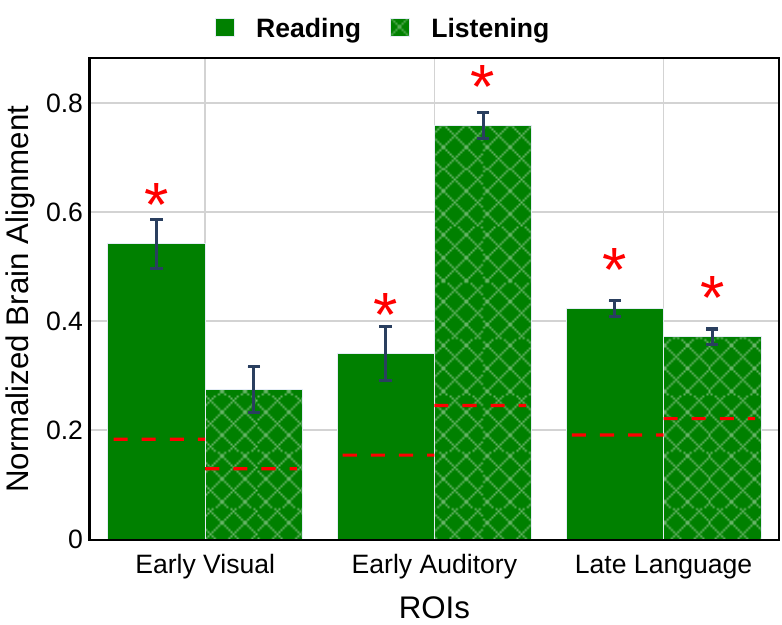}
\\(b) Speech Models
\end{minipage}
\caption{ROI-based normalized brain alignment was computed by averaging across participants, models, layers, and voxels.  \textcolor{orange}{Orange:} text,  \textcolor{green}{Green: }speech, Solid: reading, Patterned: listening. \textcolor{red}{Red} dashed line: chance performance, and \textcolor{red}{*} at a particular bar indicates that the model's prediction performance is significantly better than chance.}
\label{fig:normalized_predictivity_visual_auditory_language}
\end{figure*}

\vspace{-0.1cm}
\section{Results}
\vspace{-0.1cm}




We investigate the alignment of text- and speech-based language models during reading and listening, across early sensory processing regions (early visual and early auditory) as well as late language regions. We further use our direct approach to study how different low-level stimulus features impact this alignment. Unless otherwise specified, the results presented in the main paper reflect an average across model type, feature type, model layers, voxels, and participants, normalized by the cross-subject prediction accuracy. Error bars indicate the standard error of the mean across participants. Chance performance for each case is indicated by a \textcolor{red}{dashed} horizontal line. Results for individual models and layers are included in the appendix.

\begin{figure*}[t] 
\centering
\begin{minipage}{0.48\textwidth}
\centering
\includegraphics[width=\linewidth]{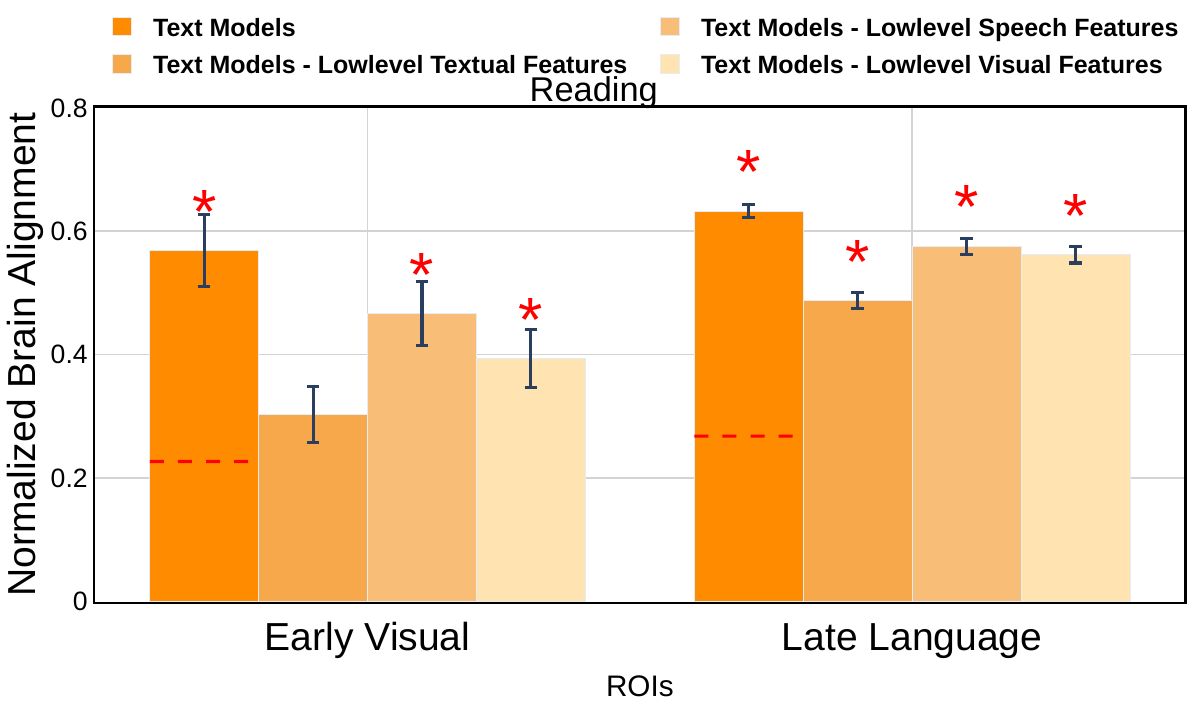}
\\(a) Text Models\\
\end{minipage}
\begin{minipage}{0.51\textwidth}
\centering
\includegraphics[width=\linewidth]{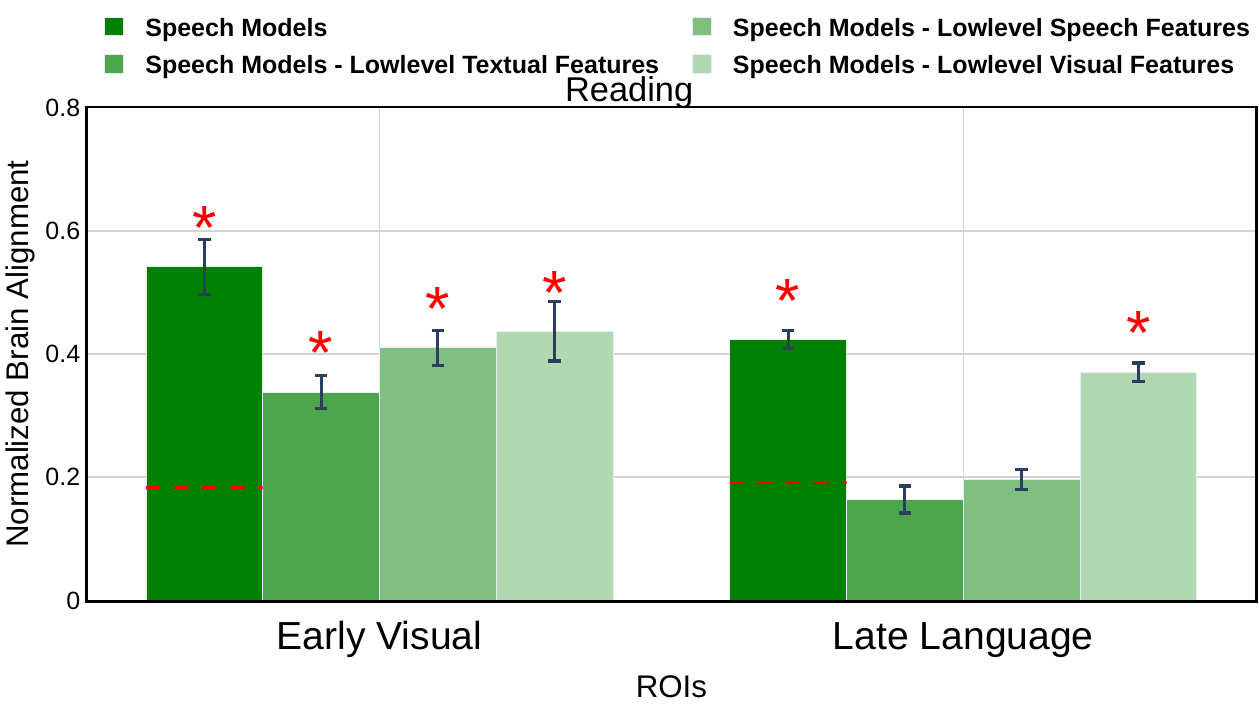}
\\(b) Speech Models\\
\end{minipage}
\caption{Reading condition in the Early Visual and Late Language Regions: \emph{(a) } For Text-based Models: Average normalized brain alignment was computed across participants before and after removal of low-level stimulus features, across layers and voxels. \emph{(b) } For Speech-based Models: Similar alignment analysis was conducted for speech models before and after removal of low-level stimulus features, across layers and voxels. \textcolor{red}{Red} dashed line: chance performance, and \textcolor{red}{*} indicates that the residuals prediction performance is significantly better than chance.}
\label{fig:normalized_predictivity_auditory_listening}
\end{figure*}

\begin{figure*}[t] 
\centering
\begin{minipage}{0.48\textwidth}
\centering
\includegraphics[width=\linewidth]{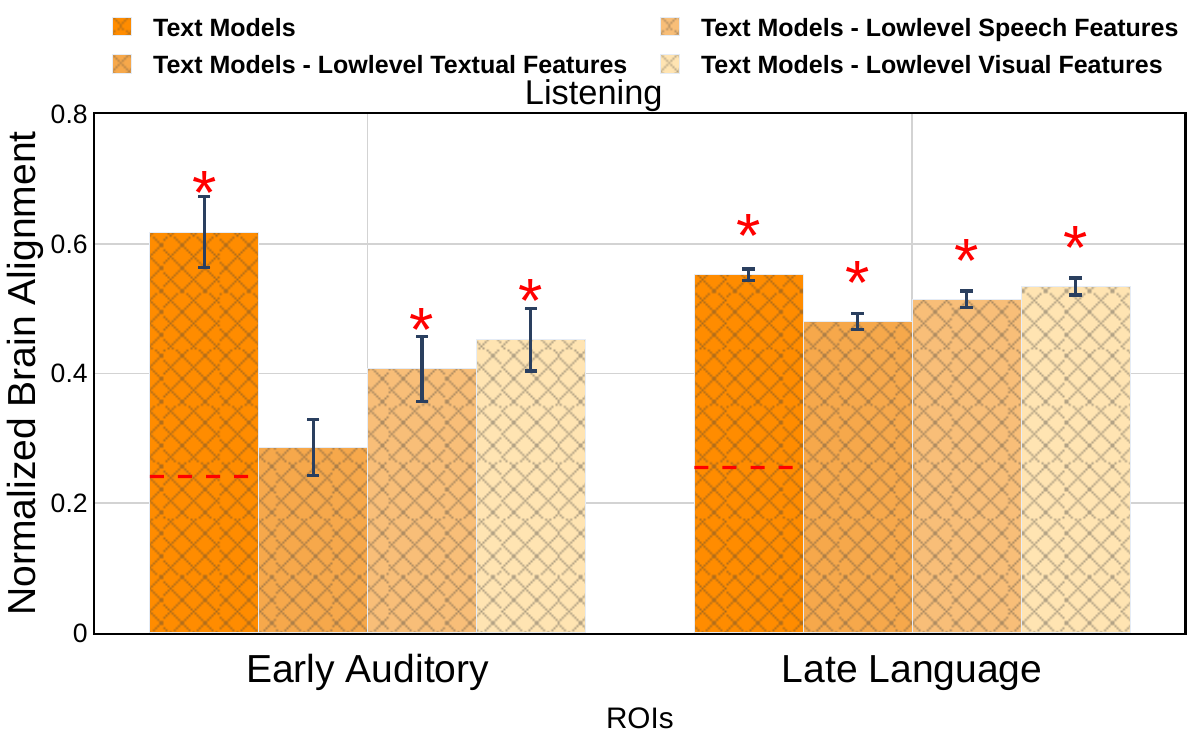}
\\(a) Text Models\\
\end{minipage}
\begin{minipage}{0.51\textwidth}
\centering
\includegraphics[width=\linewidth]{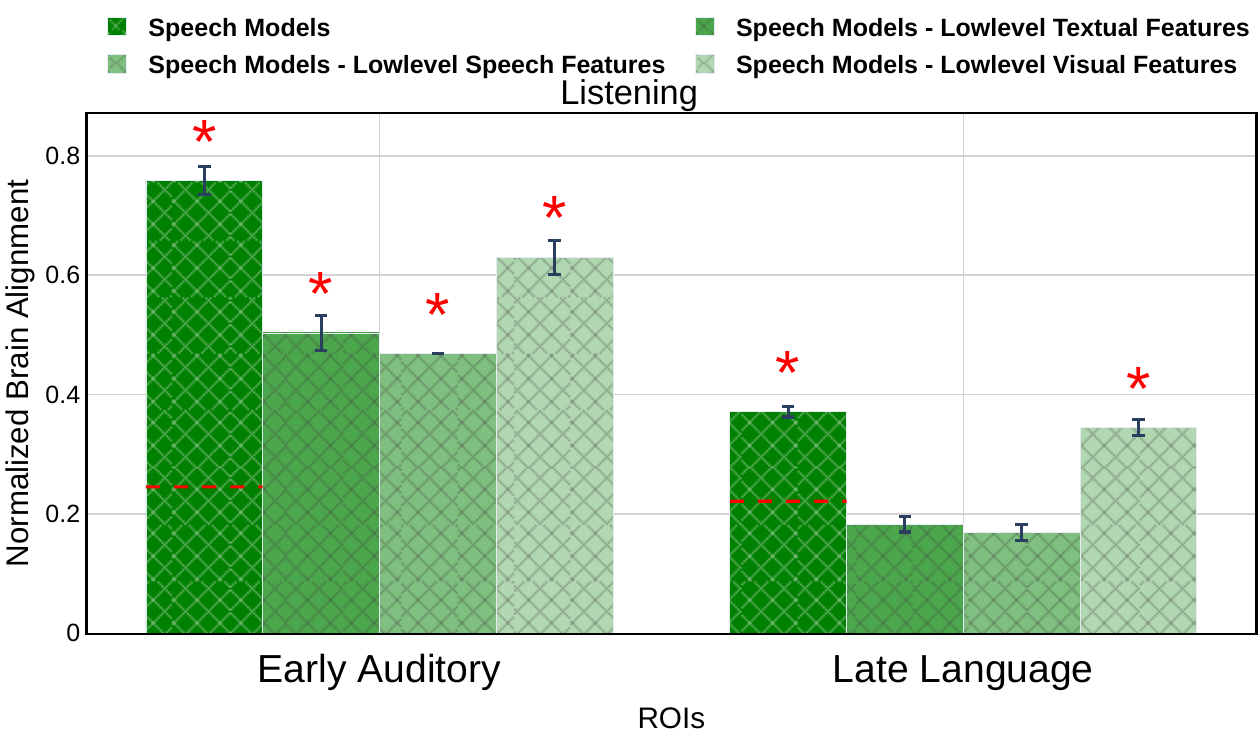}
\\(b) Speech Models\\
\end{minipage}
\caption{Listening condition in the Early Auditory and Late Language Regions: \emph{(a) } For Text-based Models: Average normalized brain alignment was computed across participants before and after removal of low-level stimulus features, across layers and voxels. \emph{(b) } For Speech-based Models: Similar alignment analysis was conducted for speech models before and after removal of low-level stimulus features, across layers and voxels. \textcolor{red}{Red} dashed line: chance performance, and \textcolor{red}{*} indicates that the residuals prediction performance is significantly better than chance.}
\label{fig:normalized_predictivity_late_language}
\end{figure*}


\begin{figure*}[t] 
\centering
\includegraphics[width=\linewidth]{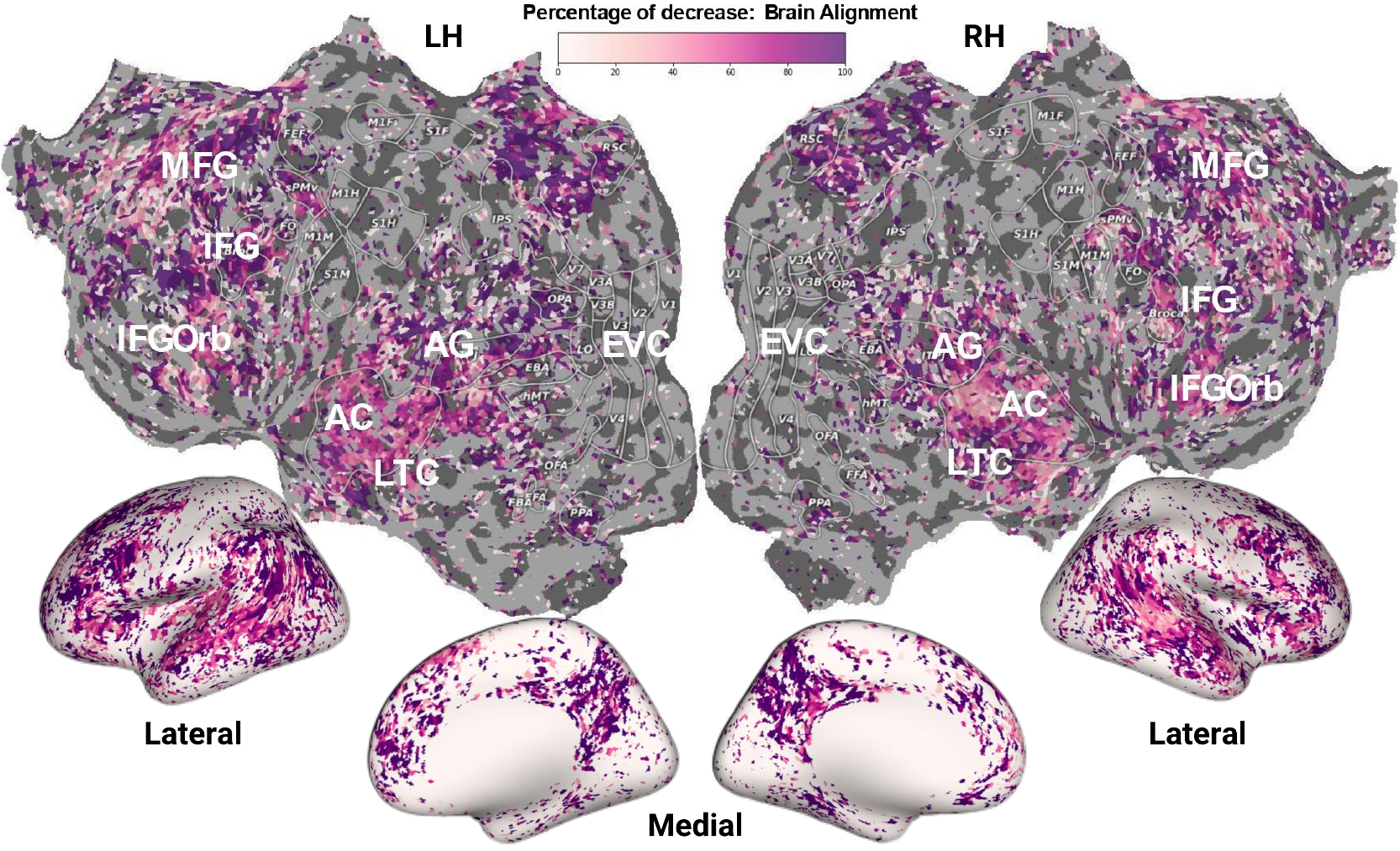}
\caption{Percentage of decrease in alignment during listening for each voxel after the removal of Phonological features from Wav2Vec2.0, projected onto the flattened cortical surface of a representative subject (subject-8). 
White indicates voxels where Phonological features do not explain any shared information within Wav2Vec2.0, and purple indicates voxels where they explain all the information predicted by Wav2Vec2.0.
}
\label{fig:2dcolormap_wav2vec_residual_phonological}
\end{figure*}

\vspace{-0.1cm}
\subsection{Brain Alignment of Text-Based Language Models}
\label{subsection1}
\vspace{-0.1cm}

In late language regions, we find that text-based language models: 
(1) have a very high brain alignment both during reading and listening (Fig.~\ref{fig:normalized_predictivity_visual_auditory_language}a), 
(2) retain most of their alignment (significantly above chance level) even after removal of low-level features both during reading (Fig.~\ref{fig:normalized_predictivity_auditory_listening}a) and listening (Fig.~\ref{fig:normalized_predictivity_late_language}a),
and (3) have slightly better alignment during reading vs. listening, however this difference largely vanishes after the removal of low-level textual features (Figs.~\ref{fig:normalized_predictivity_auditory_listening}a \&~\ref{fig:normalized_predictivity_late_language}a). These results indicate that, irrespective of the stimulus modality, the alignment of text-based language models with late language regions is not due to low-level features.

Turning to the early sensory regions, we find that the alignment in the early visual regions during reading is almost as high as it is in late language regions (Fig.~\ref{fig:normalized_predictivity_visual_auditory_language}a). This alignment is however largely explained by low-level textual features (Fig.~\ref{fig:normalized_predictivity_auditory_listening}a). The remaining alignment is low, but still significant.
The alignment in the early auditory regions during listening is also very high, and this is almost exclusively due to low-level textual features (Fig.~\ref{fig:normalized_predictivity_late_language}a). Overall, these results show that low-level textual features are the primary factor underlying the alignment of text-based models with early sensory regions, regardless of the stimulus modality.

\noindent\textbf{Impact of individual low-level features.} The most impactful low-level textual feature for text-based models is "Number of Letters" (see Fig.~\ref{fig:2dcolormap_bert_residual_numletters} in the Appendix).
Removing this feature from BERT leads to a significant drop (80-100\%) in early visual regions during reading, but only a slight drop (0-20\%) in late language regions. We also found that removing the low-level speech feature "DiPhones" from BERT significantly decreases alignment (20-40\%) even in late language regions (see Fig.~\ref{fig:2dcolormap_bert_residual_diphone} in the Appendix), presumably because short words contribute significantly to alignment provided by diphones ~\citep{gong2023phonemic} and BERT representations may contain brain-relevant information about short words.
\vspace{-0.1cm}brain
\subsection{Brain Alignment of Speech-Based Language Models}
\vspace{-0.1cm}

\label{subsection2}
In late language regions, we find that speech-based language models 
(1) have high brain alignment both during reading and listening, but not nearly as high as their text-based counterparts (Fig.~\ref{fig:normalized_predictivity_visual_auditory_language}), 
(2) lose their entire alignment (down to chance) after removal of low-level features both during reading (Fig.~\ref{fig:normalized_predictivity_auditory_listening}b) and listening (Fig.~\ref{fig:normalized_predictivity_late_language}b),
(3) have slightly better alignment during reading vs. listening, however this difference largely vanishes after the removal of low-level features (Figs.~\ref{fig:normalized_predictivity_auditory_listening}b \&~\ref{fig:normalized_predictivity_late_language}b). We further verify that the same results hold even when the input window to the speech-based language models is extended beyond 2sec to 16, 32, and 64sec (see Appendix Fig.~\ref{fig:reading_listening_normalized_predictivity_speech_windows}). These results suggest that the alignment of speech-based language models with late language regions is not due to brain-relevant semantics.

Turning to the early sensory regions, we find that the alignment in the early visual regions during reading is even higher than it is in late language regions (Fig.~\ref{fig:normalized_predictivity_visual_auditory_language}b). This alignment is partially explained by low-level textual features (Fig.~\ref{fig:normalized_predictivity_auditory_listening}b). The remaining alignment is significant and considerable.
The alignment in the early auditory regions during listening is extremely high, and this is only partially due to low-level textual or speech features. Significant unexplained alignment remains after the removal of a comprehensive set of low-level features. These results suggest that there is additional information beyond the low-level features considered in this study that is processed in the early sensory regions (Fig.~\ref{fig:normalized_predictivity_late_language}b), and captured by speech-based language models.

\noindent\textbf{Impact of individual low-level features.} The most impactful low-level feature for speech-based language models is "Phonological".
During listening, removing this feature from Wav2Vec2.0 leads to a substantial drop (80-100\%) in performance in late language regions (Fig.~\ref{fig:2dcolormap_wav2vec_residual_phonological}; see also Appendix  Fig.~\ref{fig:wav2vec_layerwise_normalized_predictivity_a1_phonological} for comprehensive layer-wise results). This indicates that speech-based models capture little, if any, brain-relevant information in these regions beyond low-level speech features.

Additionally, we observe that the "Phonological" features correlate with other low-level features (ranging from 0.62 to 0.7 Pearson correlation, see Appendix section~\ref{phonological_features_impact}; Fig.~\ref{fig:correlation_phonological}), but not as much as these features correlate among themselves. This suggests that while there is some shared variance between phonological features and number of letters, phonemes, and words, the phonological features are likely also capturing unique variance that isn't completely explained by these other features. Thus, phonological features may capture additional information over the other low-level features.

\section{Discussion and Conclusion}
Using a direct approach, we evaluated what types of information language models truly predict in brain responses. This is achieved by removing information related to specific low-level stimulus features (textual, speech, and visual) and observing how this perturbation affects the alignment with fMRI brain recordings acquired while participants read versus listened to the same naturalistic stories. 

Our direct approach leads to three key conclusions: (1) The unexpected alignment of models with sensory regions associated with the incongruent modality (i.e. text models with auditory regions and speech models with visual regions) is entirely due to low-level stimulus features that are correlated between text and speech (e.g. number of letters and number of phonemes); (2) Models exhibit varied alignment with the respective sensory regions, and this variance cannot be ascribed to low-level stimulus features alone. Text models exhibit comparable alignment with visual and auditory regions, entirely driven by low-level textual features. In contrast, speech models display significantly greater alignment with auditory regions than visual regions, and this difference cannot be entirely accounted for by the comprehensive set of low-level stimulus features studied. This underscores the capability of speech-based language models to capture additional features crucial for the early auditory cortex, hinting at their potential to enhance our understanding of this brain region; (3) While both text- and speech-based models show substantial alignment with late language regions, the impact of low-level stimulus features on text model alignment is marginal, whereas for speech-based models, alignment is entirely driven by these low-level features. Since these regions are purported to represent semantic information, this finding implies that contemporary speech-based models lack brain-relevant semantics. Furthermore, these results imply that observed similarities between speech-based models and brain recordings in the past~\citep{vaidya2022self,millet2022toward} are largely due to low-level information, which is important to take into account when interpreting the similarity between language representations in speech-based models and the brain. This disparity raises questions about the utility of speech models in modeling late language processing, contrasting with the reassuring findings for text-based models for the same regions.

Our findings also clearly show that despite the growing popularity of text-based and speech-based language models in modeling language in the brain, we are still far from a computational model of the complete information processing steps during either listening or reading. In the future, leveraging the alignment strengths of text-based models in late language regions and speech-based models in early auditory regions may lead to improved end-to-end models of listening in the brain. 

\section{Limitations}
One limitation of our approach is that the removal method we use only removes linear contributions to language model representations. While this is sufficient to remove the effect on the brain alignment, which is also modeled as a linear function, it is possible that it does not remove all information related to the specific features from the model. Another possible limitation for the interpretation of the differences between the brain alignment of text- vs speech-based models is that the models we are using have several differences beside the stimulus modality, such as the amount of their training data and objective functions. To alleviate this concern, we have tested multiple models of each type, with different objective functions and trained on different amounts of data, and showed that the results we observe generalize within the text- and speech-based model types despite these differences. Still, it is possible that some of the differences in brain alignment we observe are due to confounding differences between model types, and there is value in investigating these questions in the future with models that are controlled for architecture, objective, and training data amounts. Lastly, our work uses brain recordings that are obtained from English-speaking participants and experimental stimuli that are in English, and therefore we use models that are mostly trained on English text and speech. It is possible that the findings would differ in other languages, and this is important to study in the future.

The alignment of text-based models with the late language regions is not explained by low-level stimulus features alone. However, these regions also process high-level semantic information (e.g., discourse-level or emotion-related)~\citep{binder2011neurobiology,wehbe2014simultaneously,bookheimer2002functional}. Future work can investigate the contribution of such features to this alignment. In addition, while impressive, the current level of alignment does not reach the estimated cross-subject prediction accuracy. Inducing brain-relevant bias can be one way to enhance the alignment of these models with the human brain~\citep{schwartz2019inducing}. Overall, further research is necessary to improve both text- and speech-based language models.

\section{Ethics Statement}
All the text-based and speech-based language models used in this work are publicly available on Huggingface and free for research use.  

The fMRI dataset we use is a well known public dataset~\footnote{\url{https://gin.g-node.org/denizenslab/narratives_reading_listening_fmri}} that has been previously used in many publications in both ML and neuroscience venues (ML: ~\citep{jain2018incorporating,jat2020relating,antonello2021low,deniz2019representation,chen2024cortical}, Neuroscience: ~\citep{huth2022gallant,deniz2019representation}. 
The dataset consists of 11 stories, 6 participants, and all participants listened to and read all the stories. The speech stimuli consisted of 10- to 15 min stories taken from The Moth Radio Hour and used previously~\citep{huth2016natural}. 

The datasets are licensed under Creative Commons CC0 1.0 Public Domain Dedication. No anticipated risks are associated with using the data from these provided links.

\section*{Acknowledgements}
This work was partially funded by the German Research Foundation (DFG) - DFG Research Unit
FOR 5368. 

\bibliography{iclr2024_conference}



\newpage
\appendix

\begin{center}
    \LARGE{\textbf{{\textit{Appendix for: Speech language models lack important brain-relevant semantics}}}}
\end{center}
\vspace{18pt}
\hrule
\vspace{18pt}

\section{Naturalistic Reading-Listeing fMRI Dataset}
\label{naturalistic_dataset}

We use the publicly available naturalistic story reading and listening fMRI dataset provided by~\citep{deniz2019representation}. The dataset consists of 11 stories, 6 participants, and all participants listened to and read all the stories. The speech stimuli consisted of 10- to 15 min stories taken from The Moth Radio Hour and used previously~\citep{huth2016natural}. The 10 selected stories cover a wide range of topics and are highly engaging. The model validation dataset consisted of one 10 min story. All stimuli were played at 44.1 kHz using the pygame library in Python. The audio of each story was down-sampled to 11.5 kHz and the Penn Phonetics Lab Forced Aligner was used to automatically align the audio to the transcript. Finally the aligned transcripts were converted into separate word and phoneme representations using Praat's TextGrid object. The word representation of each story is a list of pairs (W, t), where W is a word and t is the onset time in seconds.

\begin{figure*}[!ht] 
\centering
\includegraphics[width=0.49\linewidth]{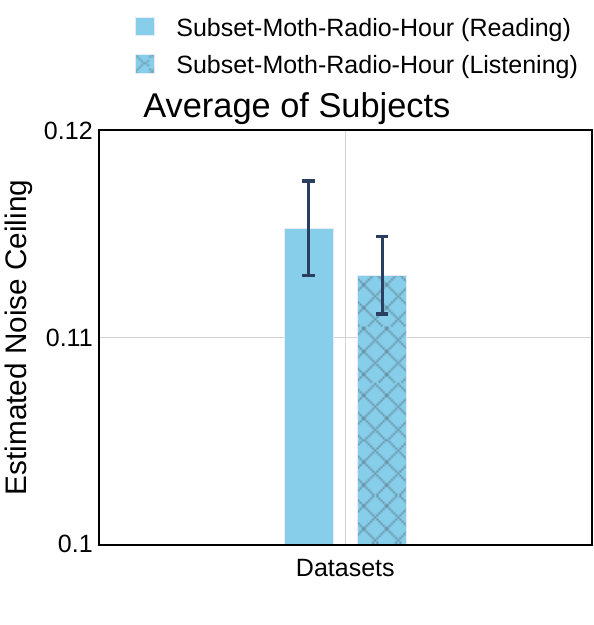}
\includegraphics[width=0.49\linewidth]{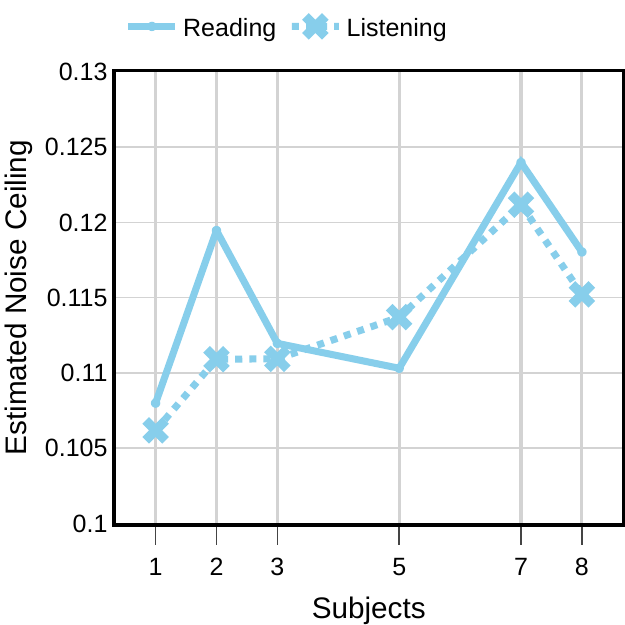}
\caption{The estimated cross-subject prediction accuracy was computed across all participants for the Subset-Moth-Radio-Hour naturalistic reading-listening fMRI dataset. The average cross-subject prediction accuracy is shown across predicted voxels where each voxel ceiling value is $\ge$ 0.05.}
\label{fig:noise_ceiling_datasets}
\end{figure*}

\begin{figure*}[!ht] 
\centering
\includegraphics[width=0.49\linewidth]{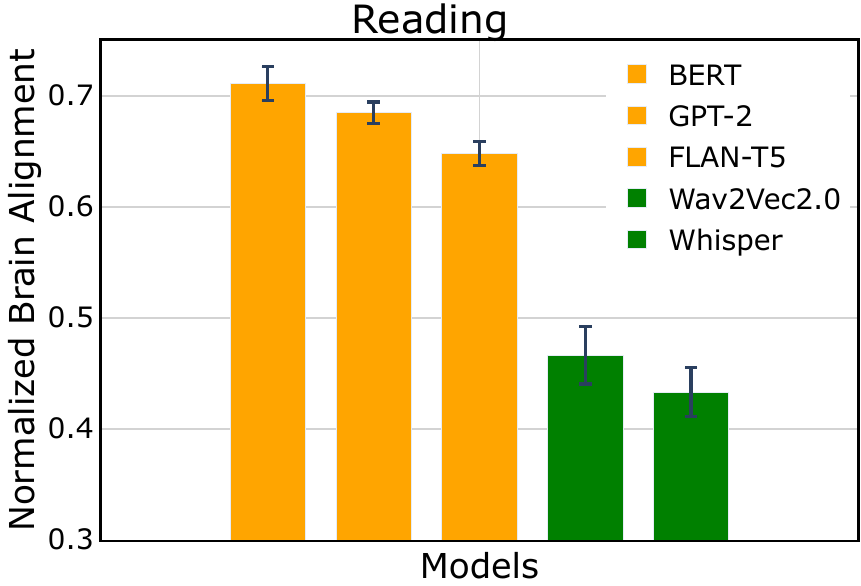}
\includegraphics[width=0.49\linewidth]{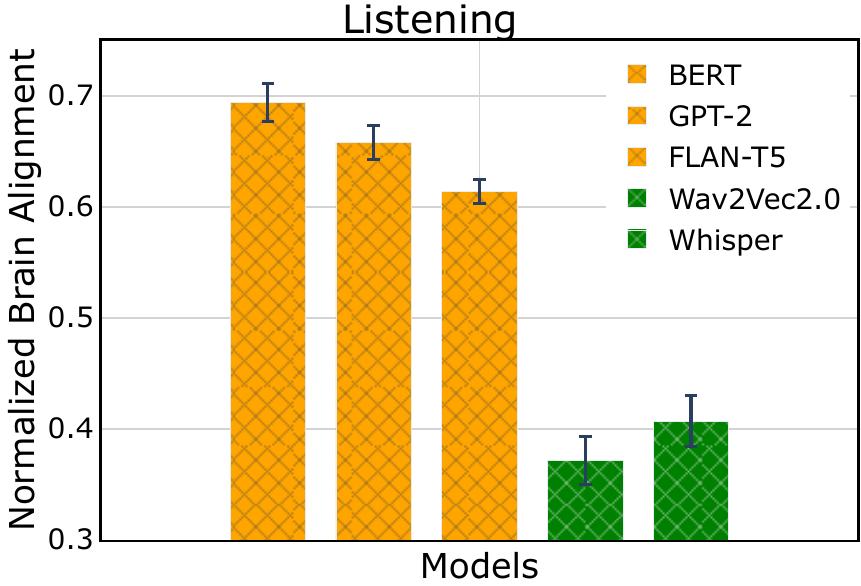}
\caption{Average normalized brain alignment was computed over the average of subjects for each model (3 text-based and 2 speech-based language models), across layers, for two modalities: reading (left) and listening (right).}
\label{fig:normalized_predictivity_datasets}
\end{figure*}

The same stories from listening sessions were used for reading sessions. Praat's word representation for each story (W, t) was used for generating the reading stimuli. The words of each story were presented one-by-one at the center of the screen using a rapid serial visual presentation (RSVP) procedure. During reading, each word was presented for a duration precisely equal to the duration of that word in the spoken story. RSVP reading is different from natural reading because during RSVP the reader has no control over which word to read at each point in time. Therefore, to make listening and reading more comparable, the authors matched the timing of the words presented during RSVP to the rate at which the words occurred during listening. This implies that the downsampling procedure remains the same for both reading and listening conditions

The total number of words in each story as follows:
Story1: 2174;
Story2: 1469;
Story3: 1964;
Story4: 1893;
Story5: 2209;
Story6: 2786;
Story7: 3218;
Story8: 2675;
Story9: 1868;
Story10: 1641;
Story11: 1839 (test dataset)

\section{Estimated Cross-Subject Prediction Accuracy}
\label{noiseceiling}

\begin{figure*}[!ht] 
\centering
\begin{minipage}{\textwidth}
\centering
    \includegraphics[width=0.81\linewidth]{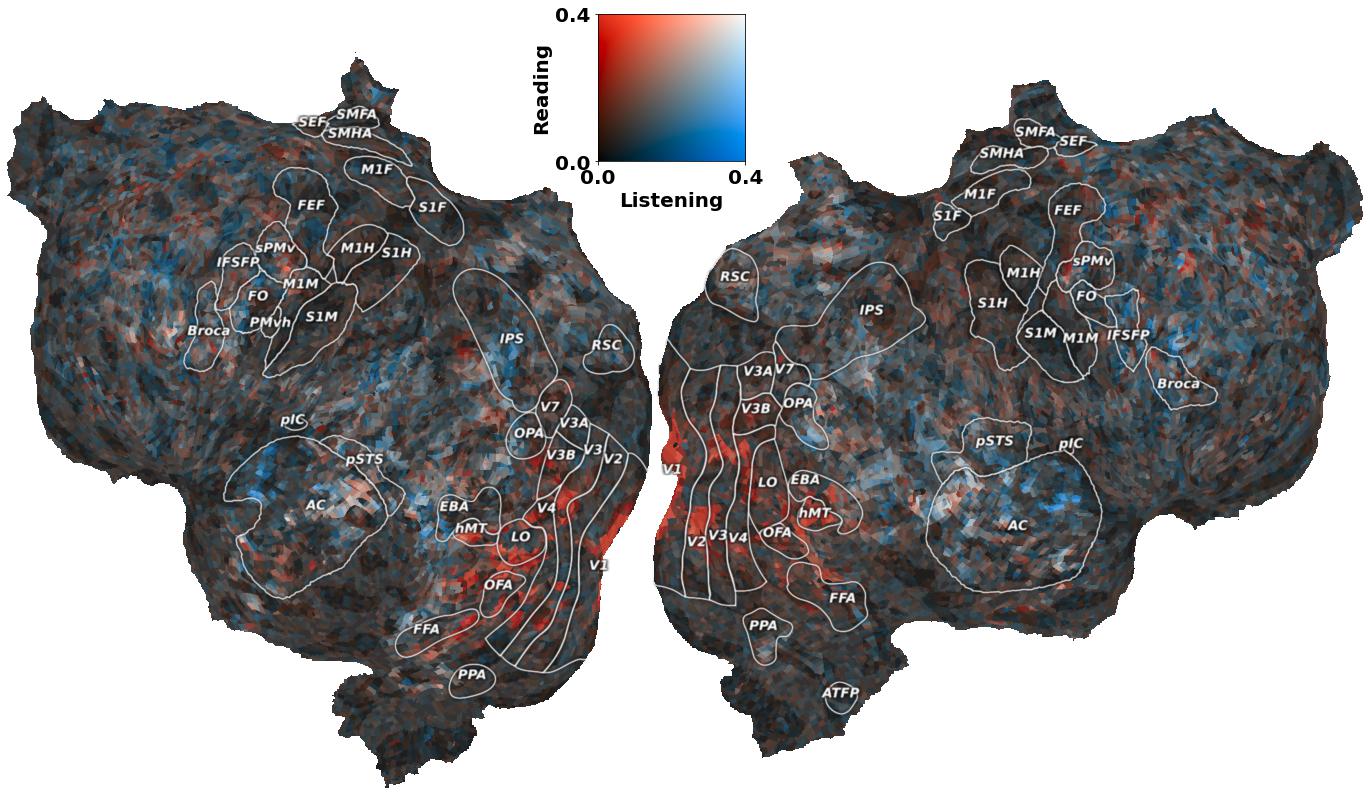}
    \\(a) Subject-01 \\
\end{minipage}
\begin{minipage}{\textwidth}
\centering
    \includegraphics[width=0.81\linewidth]{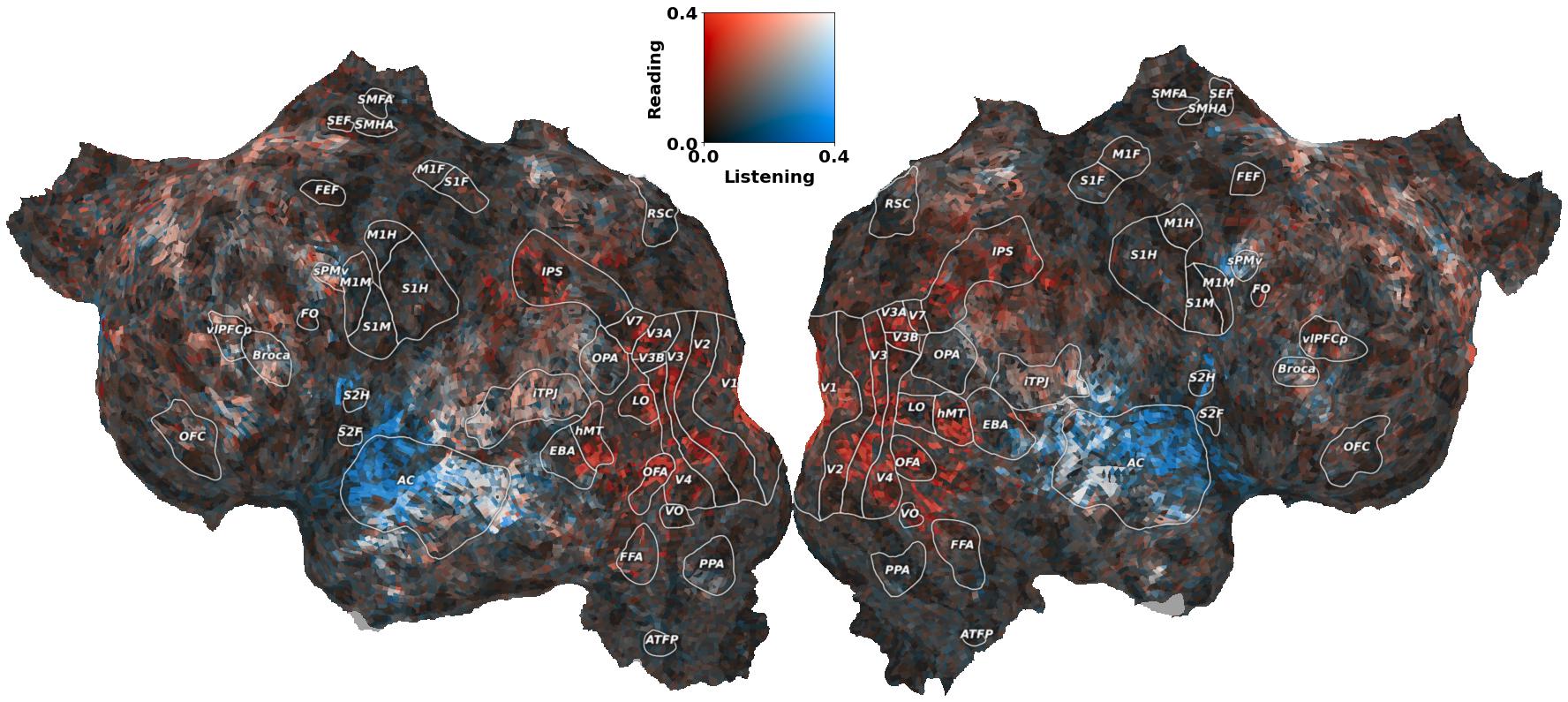}
    \\(a) Subject-02 \\
\end{minipage}
\begin{minipage}{\textwidth}
\centering
    \includegraphics[width=0.81\linewidth]{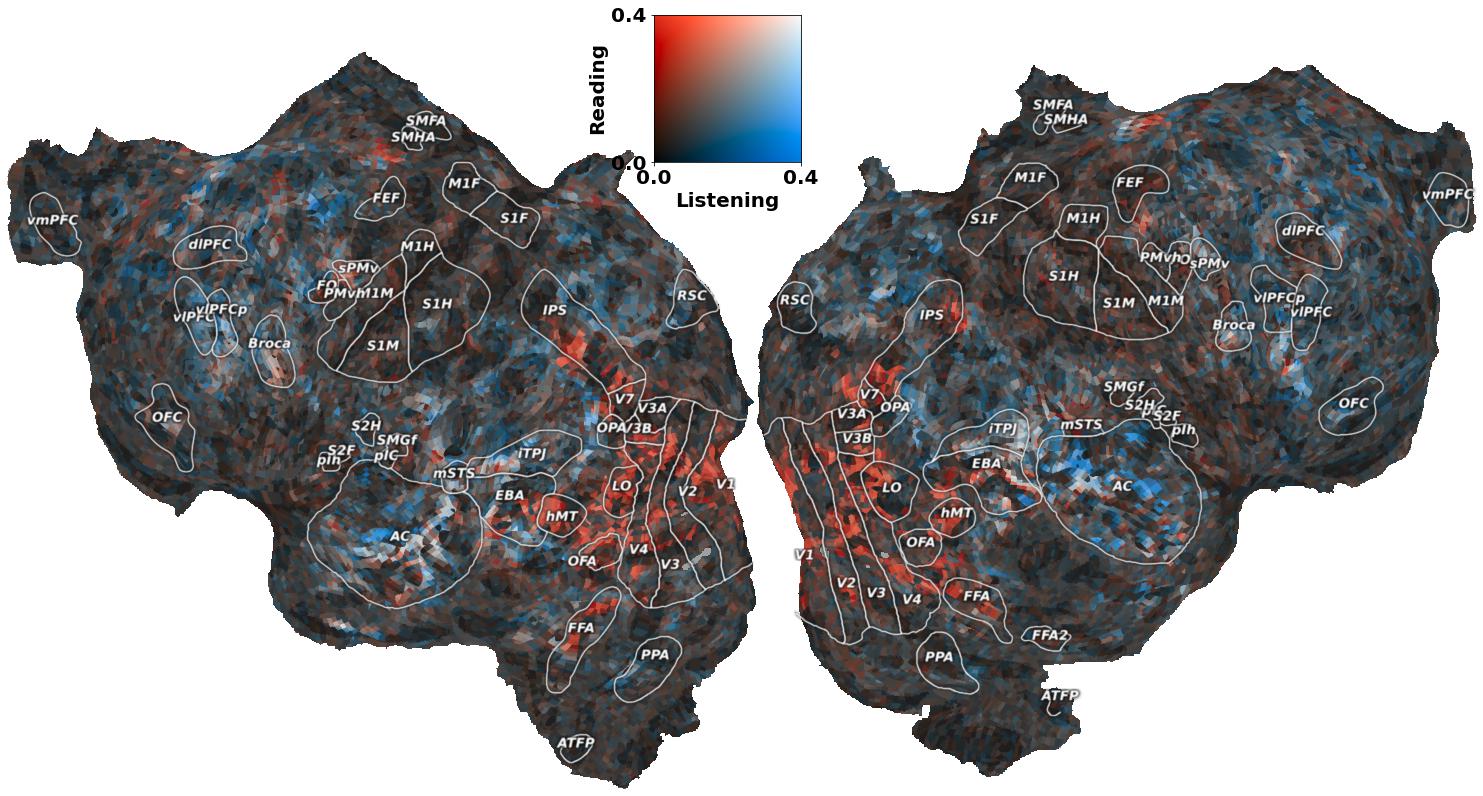}
    \\(a) Subject-03 \\
\end{minipage}
\caption{Contrast of estimated cross-subject prediction accuracy for the remaining participants for the reading vs listening condition. 
\textcolor{cyan}{BLUE-AC (Auditory Cortex)} voxels have a higher cross-subject prediction accuracy in listening, and \textcolor{red}{Red-VC (Visual Cortex)} voxels have a higher cross-subject prediction accuracy in reading. Voxels that appear in white have similar cross-subject prediction accuracy across conditions, and are distributed across language regions.}
\label{fig:noise_ceiling_subject01}
\end{figure*}

\begin{figure*}[!ht] 
\centering
\begin{minipage}{\textwidth}
\centering
    \includegraphics[width=\linewidth]{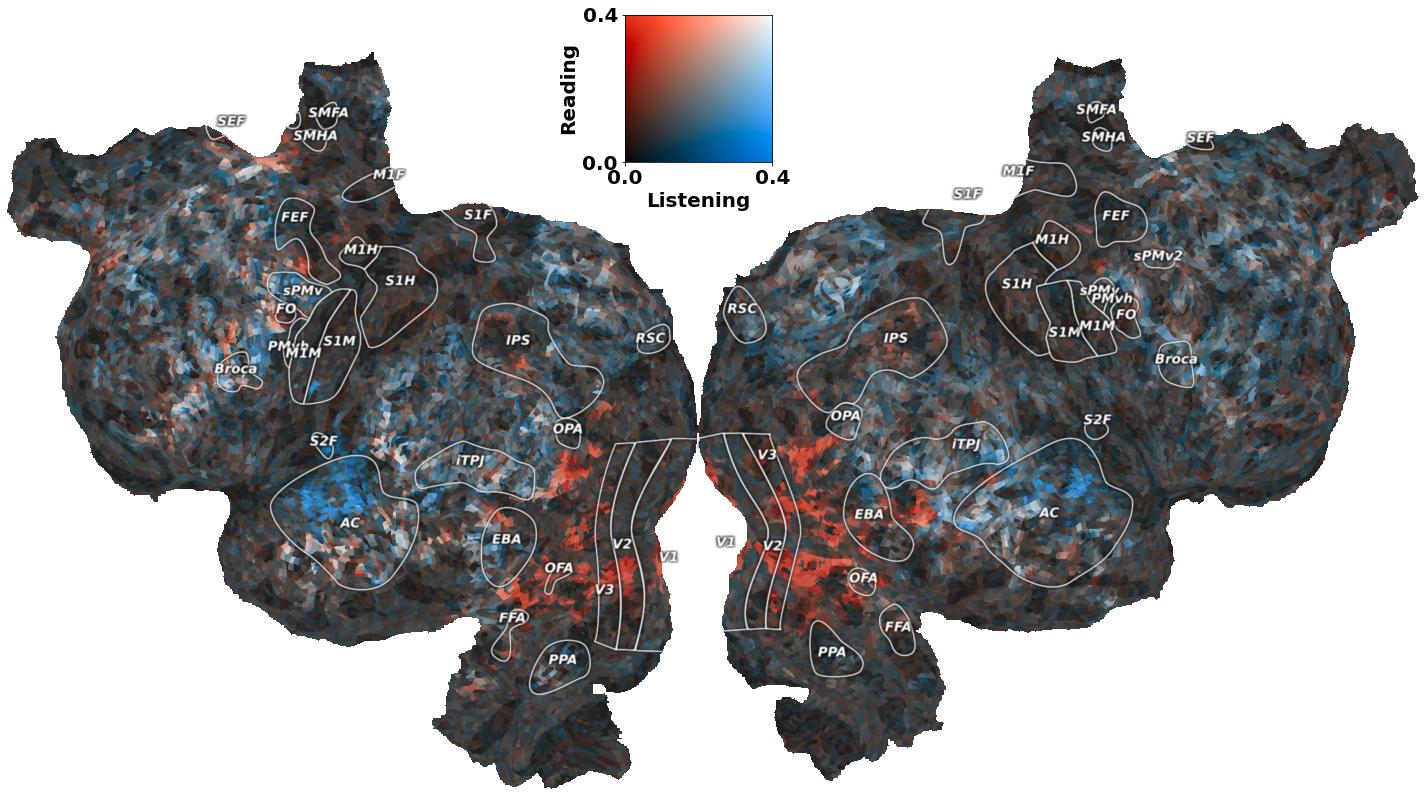}
    \\(a) Subject-05 \\
\end{minipage}
\begin{minipage}{\textwidth}
\centering
    \includegraphics[width=\linewidth]{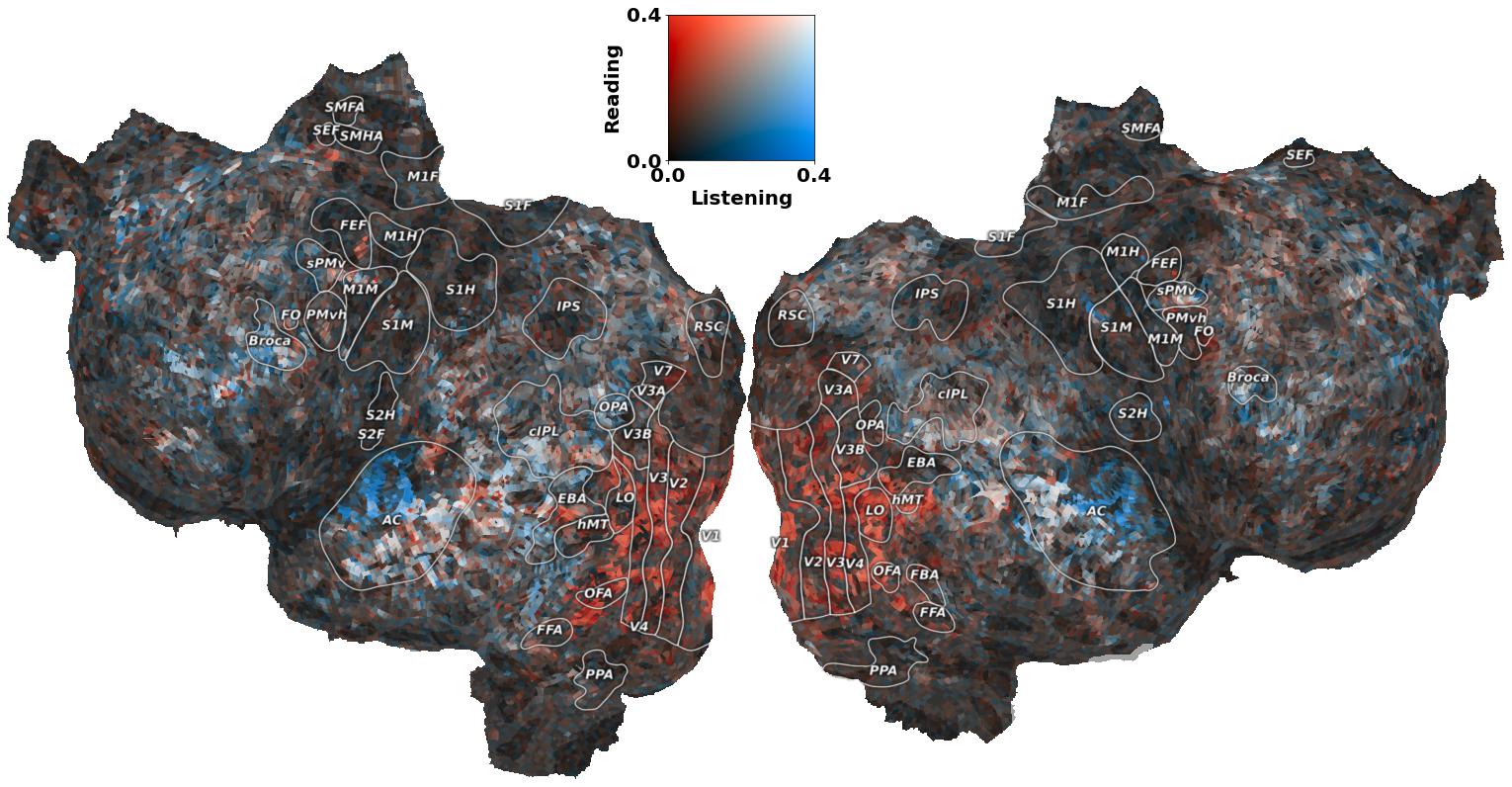}
    \\(a) Subject-07 \\
\end{minipage}
\caption{Contrast of estimated cross-subject prediction accuracy for the remaining participants for the reading vs listening condition. 
\textcolor{cyan}{BLUE-AC (Auditory Cortex)} voxels have a higher cross-subject prediction accuracy in listening, and \textcolor{red}{Red-VC (Visual Cortex)} voxels have a higher cross-subject prediction accuracy in reading. Voxels that appear in white have similar cross-subject prediction accuracy across conditions, and are distributed across language regions.}
\label{fig:noise_ceiling_subject07}
\end{figure*}

We present the average estimated cross-subject prediction accuracy across voxels for the \emph{naturalistic reading-listening fMRI} dataset in Fig.~\ref{fig:noise_ceiling_datasets}.
We observe that the average estimated cross-subject prediction accuracy across voxels for the two modalities is not significantly different. However, as depicted in Figs.~\ref{fig:noise_ceiling_subject01} and~\ref{fig:noise_ceiling_subject07}, there are clear regional differences across all the participants: Early visual regions have higher cross-subject prediction accuracy during the reading condition (\textcolor{red}{Red} voxels), while many of the early auditory regions have a higher cross-subject prediction accuracy during the listening condition (\textcolor{blue}{Blue} voxels).


\section{Whole Brain Analysis: Text vs. Speech model alignment during reading vs. listening} 
In Fig.~\ref{fig:normalized_predictivity_datasets}, we report the whole brain alignment of each model normalized by the cross-subject prediction accuracy for the naturalistic reading and listening dataset. We show the average normalized brain alignment across subjects, layers, and voxels. 
We perform the \emph{Wilcoxon signed-rank} test to test whether the differences between text and speech-based language models are statistically significant. We found that all text-based models are statistically significantly better at predicting brain responses than all speech-based models in both modalities.

\section{Dissecting Brain Alignment}
\begin{figure*}[!ht] 
\centering
\begin{minipage}{0.49\textwidth}
\centering
\includegraphics[width=\linewidth]{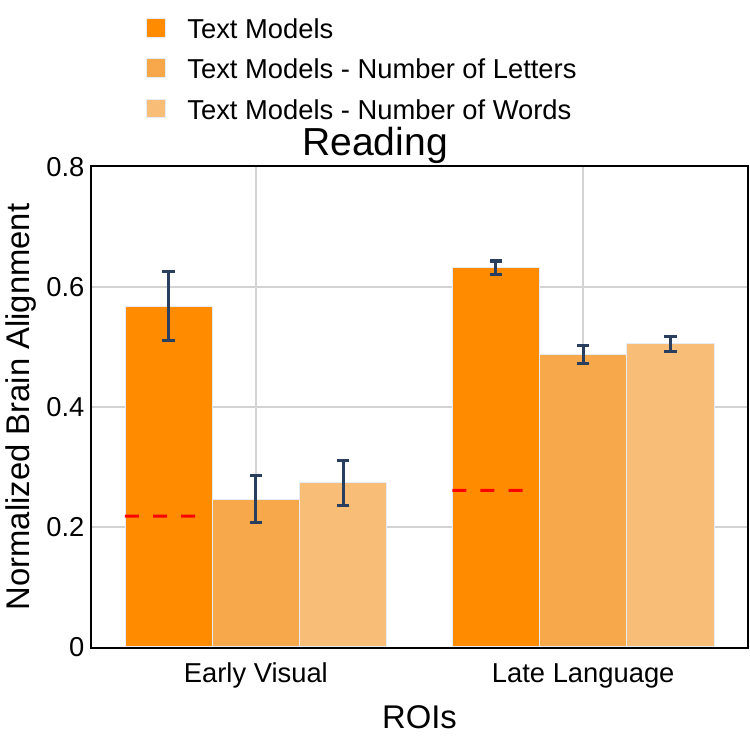}
\\(a) Removal of Low-level Textual Features\\
\end{minipage}
\begin{minipage}{0.49\textwidth}
\centering
\includegraphics[width=\linewidth]{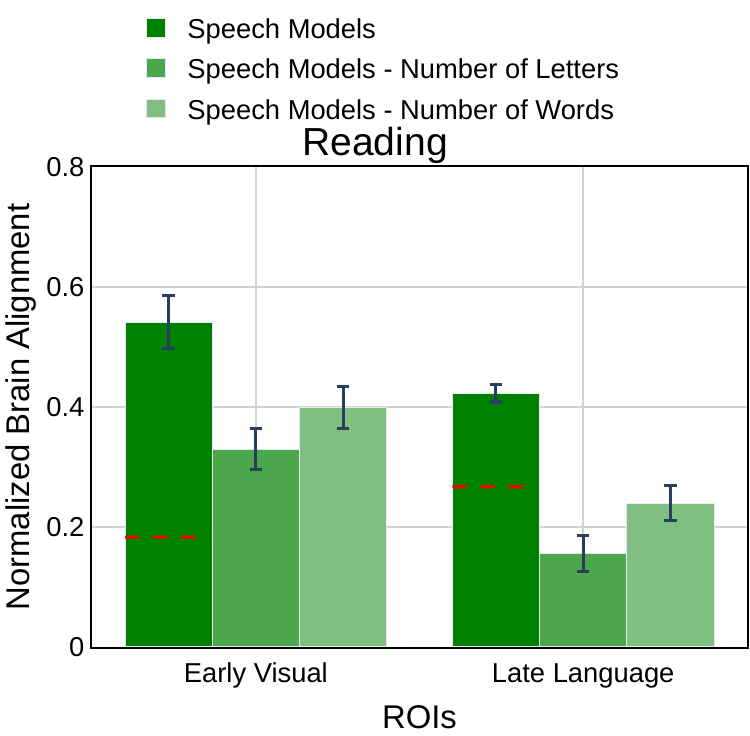}
\\(b) Removal of Low-level Textual Features\\
\end{minipage}
\begin{minipage}{0.48\textwidth}
\centering
\includegraphics[width=\linewidth]{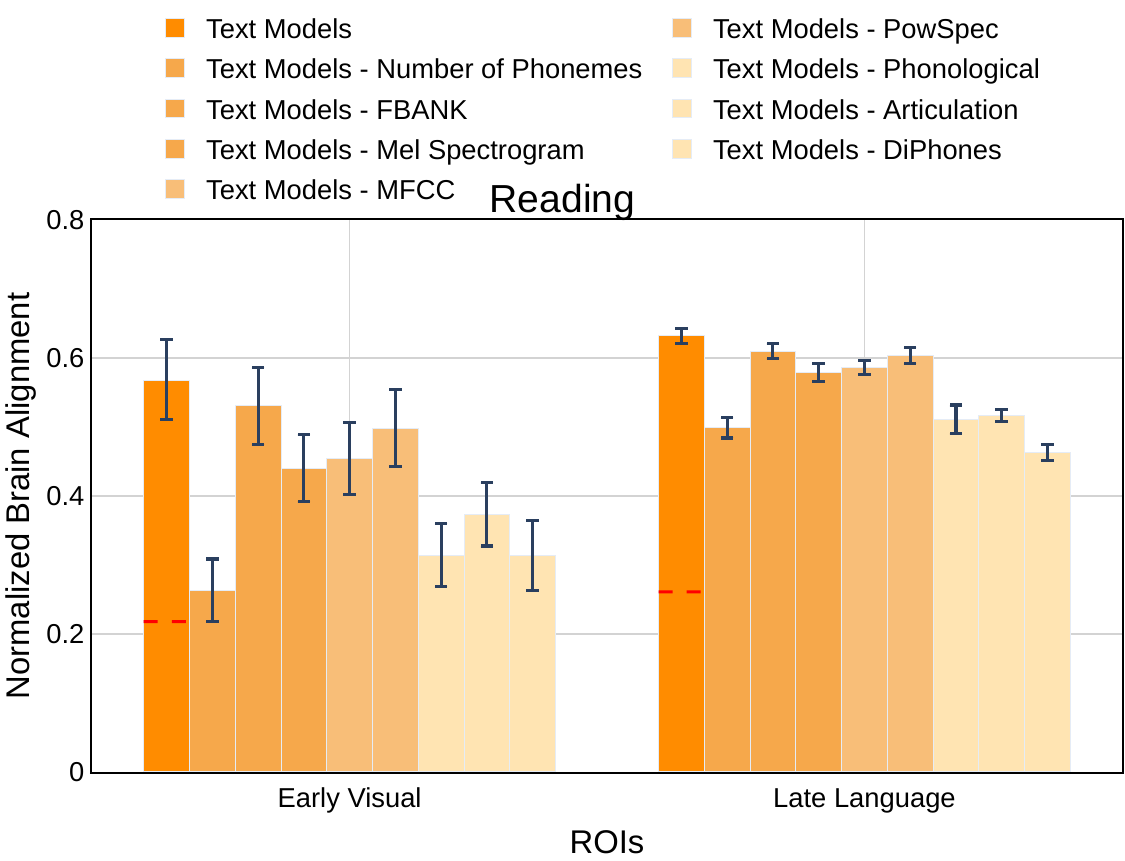}
\\(c) Removal of Low-level Speech Features\\
\end{minipage}
\begin{minipage}{0.50\textwidth}
\centering
\includegraphics[width=\linewidth]{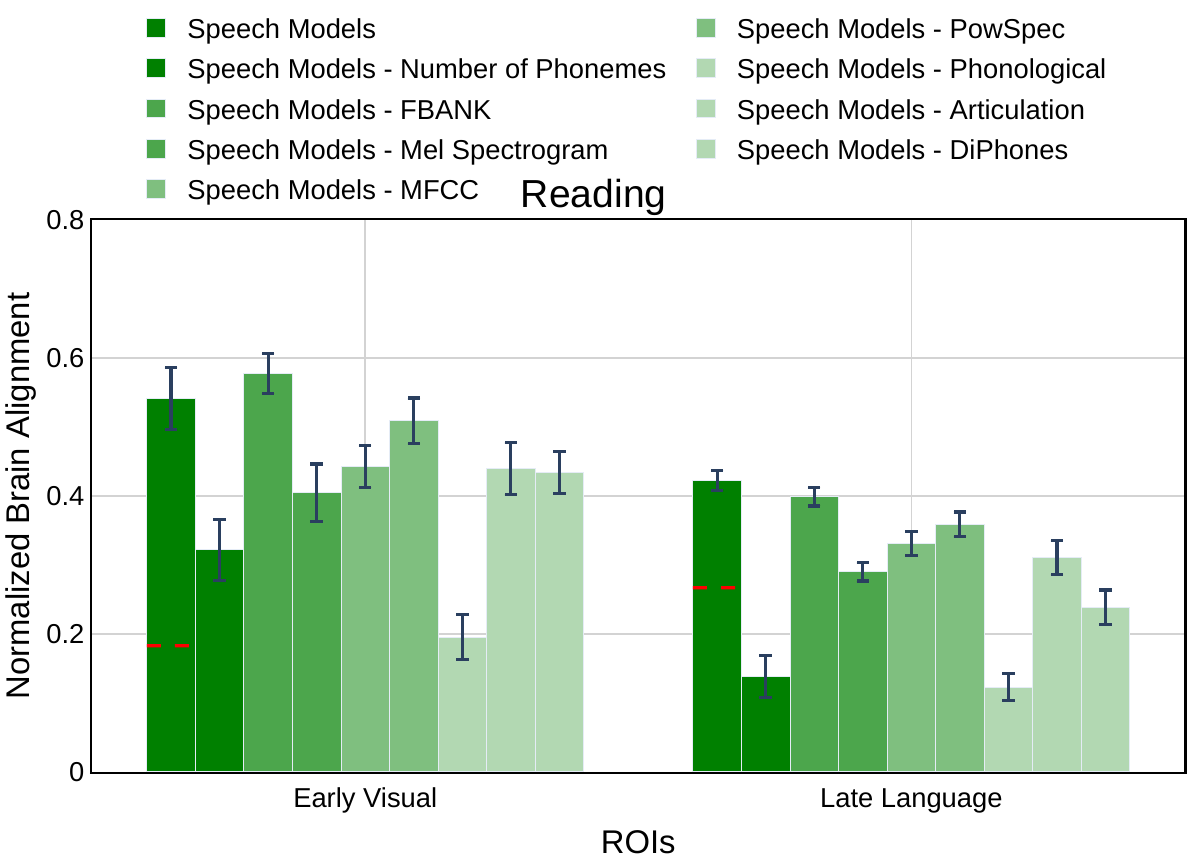}
\\(d) Removal of Low-level Speech Features\\
\end{minipage}
\caption{Brain Reading: \emph{(a) \& (b) } Removal low-level textual features, Average normalized brain alignment was computed over the average of participants for text and speech-based models, across layers for each low-level textual property. \emph{(c) \& (d) } Removal of low-level speech features: Average normalized brain alignment was computed across participants for text and speech-based models, across layers for each low-level speech property.}
\label{fig:normalized_predictivity_reading_properties}
\end{figure*}

Our major goal of the current study is to identify the specific types of information these language models capture in brain responses. To achieve this, we remove information related to
specific low-level stimulus features (textual, speech, and visual) in the language model representations, and observe how this perturbation affects the alignment with fMRI brain recordings acquired while participants read versus listened to the same naturalistic stories.
In subsections~\ref{subsection1} and~\ref{subsection2} (see in the main paper), all our results presented are averaged within types of models and types of low-level stimulus feature categories.
Here, we report the residual performance results of individual low-level stimulus features for both text- and speech-based language models, as shown in Figs.~\ref{fig:normalized_predictivity_listening_properties} and~\ref{fig:normalized_predictivity_reading_properties}.

\subsection{Why do text-based language models predict speech-evoked brain activity in early auditory cortices?}
\begin{figure*}[!ht] 
\centering
\begin{minipage}{0.49\textwidth}
\centering
\includegraphics[width=\linewidth]{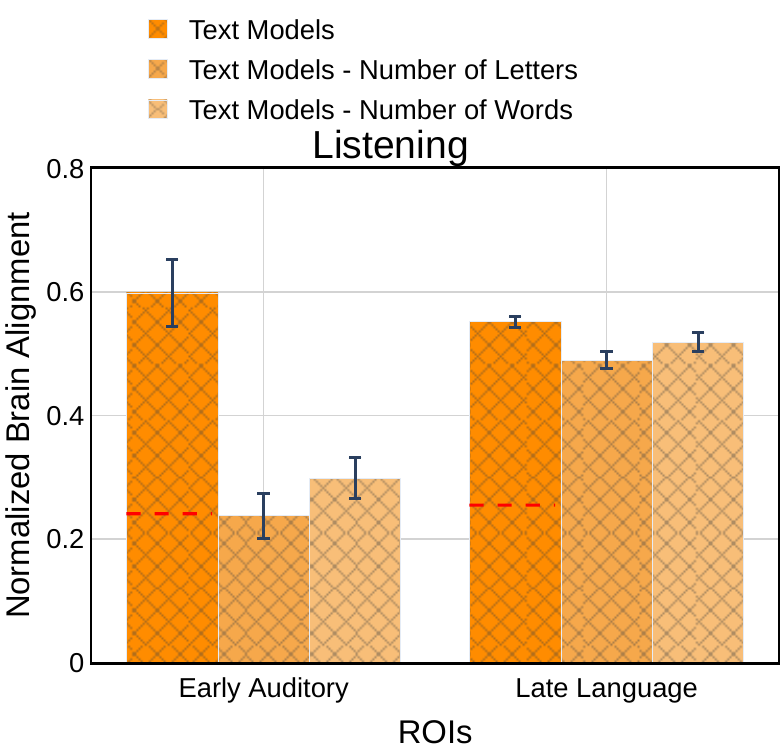}
\\(a) Removal of Low-level Textual Features\\
\end{minipage}
\begin{minipage}{0.49\textwidth}
\centering
\includegraphics[width=\linewidth]{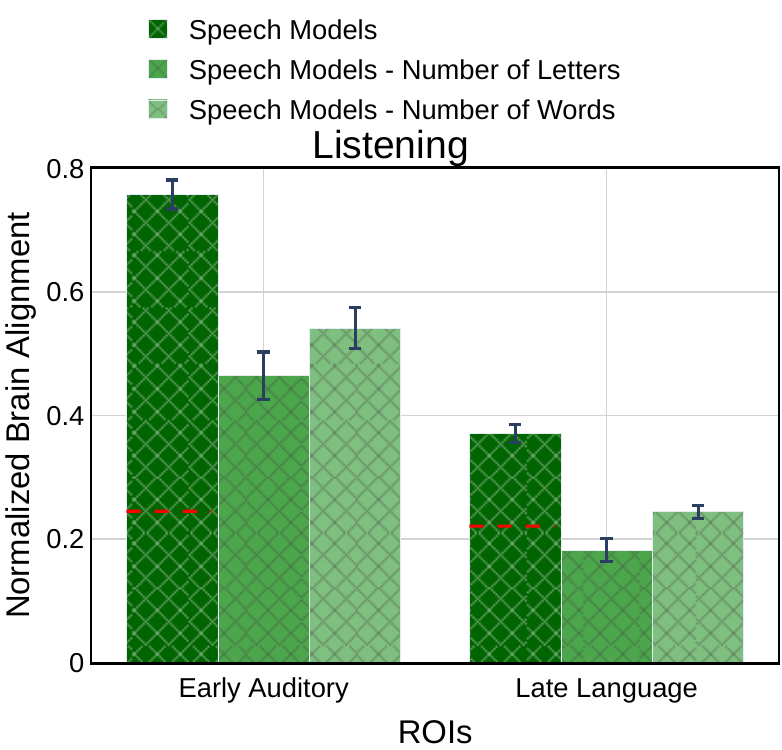}
\\(b) Removal of Low-level Textual Features\\
\end{minipage}
\begin{minipage}{0.48\textwidth}
\centering
\includegraphics[width=\linewidth]{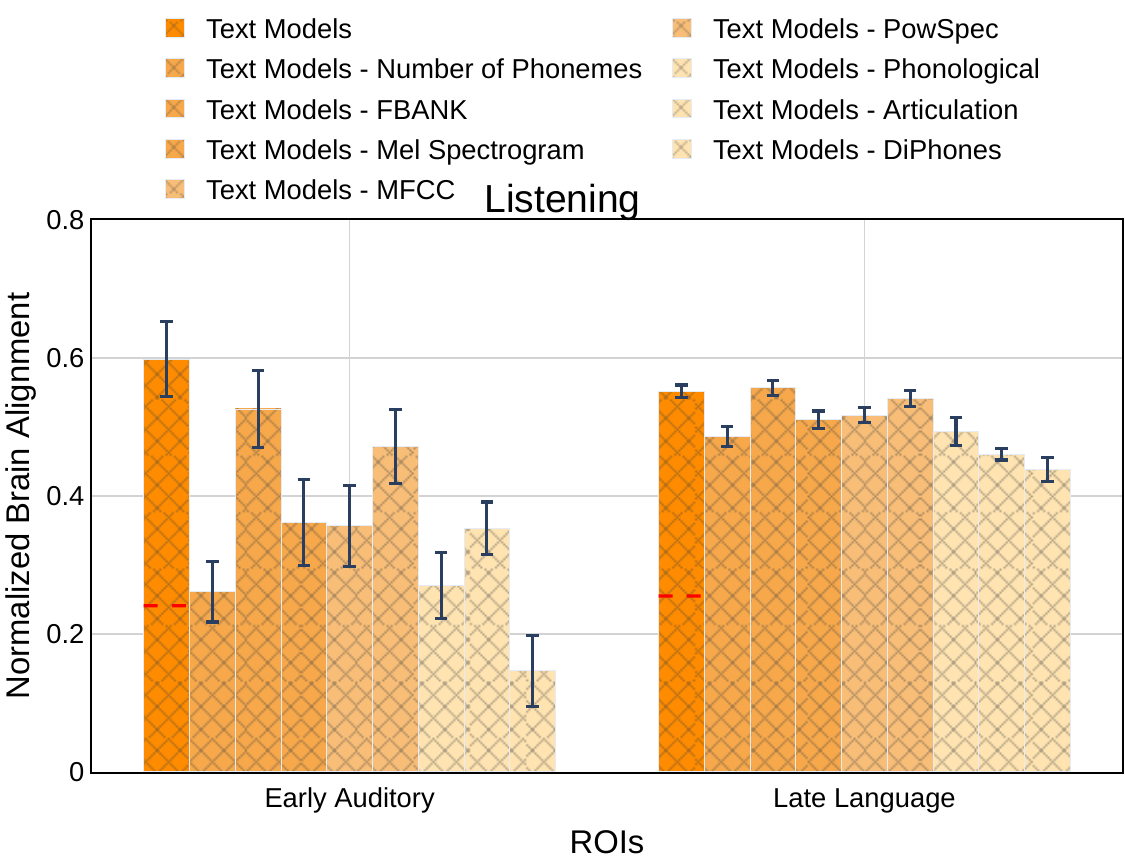}
\\(c) Removal of Low-level Speech Features\\
\end{minipage}
\begin{minipage}{0.50\textwidth}
\centering
\includegraphics[width=\linewidth]{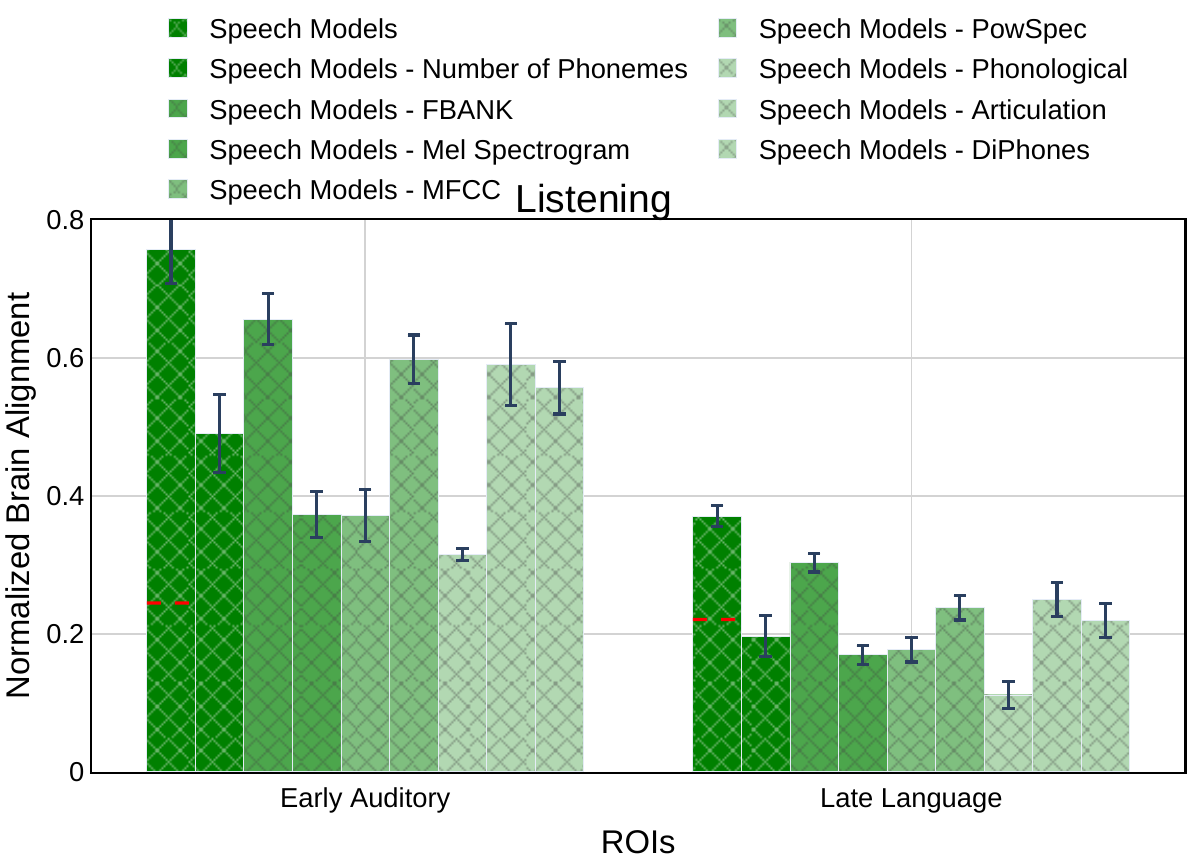}
\\(d) Removal of Low-level Speech Features\\
\end{minipage}
\caption{Brain Listening: \emph{(a) \& (b) } Removal of Low-level textual features, Average normalized brain alignment was computed over the average of participants for text and speech-based models, across layers for each low-level textual property. \emph{(c) \& (d) } Removal of low-level speech features: Average normalized brain alignment was computed across participants for text and speech-based models, across layers for each low-level speech property.}
\label{fig:normalized_predictivity_listening_properties}
\end{figure*}

In Fig.~\ref{fig:normalized_predictivity_listening_properties}, we report the normalized brain alignment results during listening in the early auditory cortex for both text- and speech-based language models, along with their residual performance after eliminating low-level stimulus features.

\noindent\textbf{Removal of low-level textual features}
We make the following observations from Fig.~\ref{fig:normalized_predictivity_listening_properties} (a) \& (b): (1) Removal of number of letters feature results in a larger performance drop (more than 30\% of the original performance) for both text- and speech-based language models.
(2) Similarly removal of number of words feature also affect more than 25\% drop indicate that low-level textual features are captured in both text and speech-based language models.

\noindent\textbf{Removal of low-level speech features}
We make the following observations from Fig.~\ref{fig:normalized_predictivity_listening_properties} (c) \& (d):
(1) Removal of phonological features results in a larger performance drop (more than 50\% of the original performance) for speech-based language models than text-based models (30\% drop of the original performance).
(2) Additionally, the removal of low-level speech features such as Mel spectrogram, MFCC and DiPhones leads to major performance drop (more than 40\%) for speech-based language models.
(3) In contrast, the removal of remaining low-level speech features, including FBANK, PowSpec and Articulation, has less shared information with speech-based language models and results in a minor performance drop (i.e. less than 20\%).
These findings indicate that speech-based language models outperform text-based language models because they better leverage low-level speech features such as MFCC, Mel spectrogram, and Phonological.
Overall, phonological features are the largest contributors for for both text and speech-based language models.
Specifically, the presence of correlated information in phonological features related to low-level textual (e.g., number of letters) and speech (e.g., number of phonemes) features explains a large portion of the brain alignment for both types of models. 


\subsection{Why do both types of models exhibit similar degree of brain alignment in early visual cortices?}

In Fig.~\ref{fig:normalized_predictivity_reading_properties}, we report the normalized brain alignment results during reading in the early visual cortex for both text- and speech-based language models, along with their residual performance after eliminating low-level stimulus features.

\noindent\textbf{Removal of low-level textual features}
We make the following observations from Fig.~\ref{fig:normalized_predictivity_reading_properties} (a) \& (b): (1) Similar to the listening condition in the early auditory regions, the removal of number of letters feature from both types of models leads to a significant drop in brain alignment in the early visual region. 
(2) Furthermore, the removal of number of words feature also leads to a drop of more than 20\% in the early visual region indicate that low-level textual features are captured in both text and speech-based language models.
This indicates that the performance of both types of models in early visual cortices is largely due to the number of letters feature followed by the number of words. 

\noindent\textbf{Removal of low-level speech features}
We make the following observations from Fig.~\ref{fig:normalized_predictivity_reading_properties} (c) \& (d):
(1) In the early visual region, removal of phonological features results in a larger performance drop (more than 35\% of the original performance) for speech-based language models than text-based models (20\% drop of the original performance).
(2) However, the removal of remaining low-level speech features has less shared information with text-based language models and results in a minor performance drop (i.e. less than 10\%).
(3) In the visual word form area, the removal of all low-level speech features has no affect on brain alignment for text-based language models, while the removal of phonological features from speech-based models results in alignment dropping to zero.
Overall, phonological features are the largest contributors for speech-based language models, both in early visual and visual word form areas. 

\begin{figure*}[!ht] 
\centering
\includegraphics[width=0.8\linewidth]{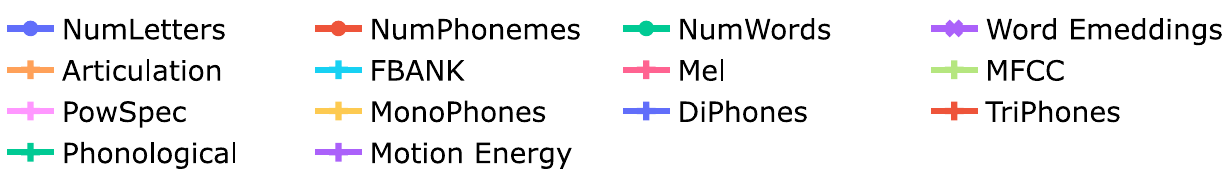}
\includegraphics[width=0.47\linewidth]{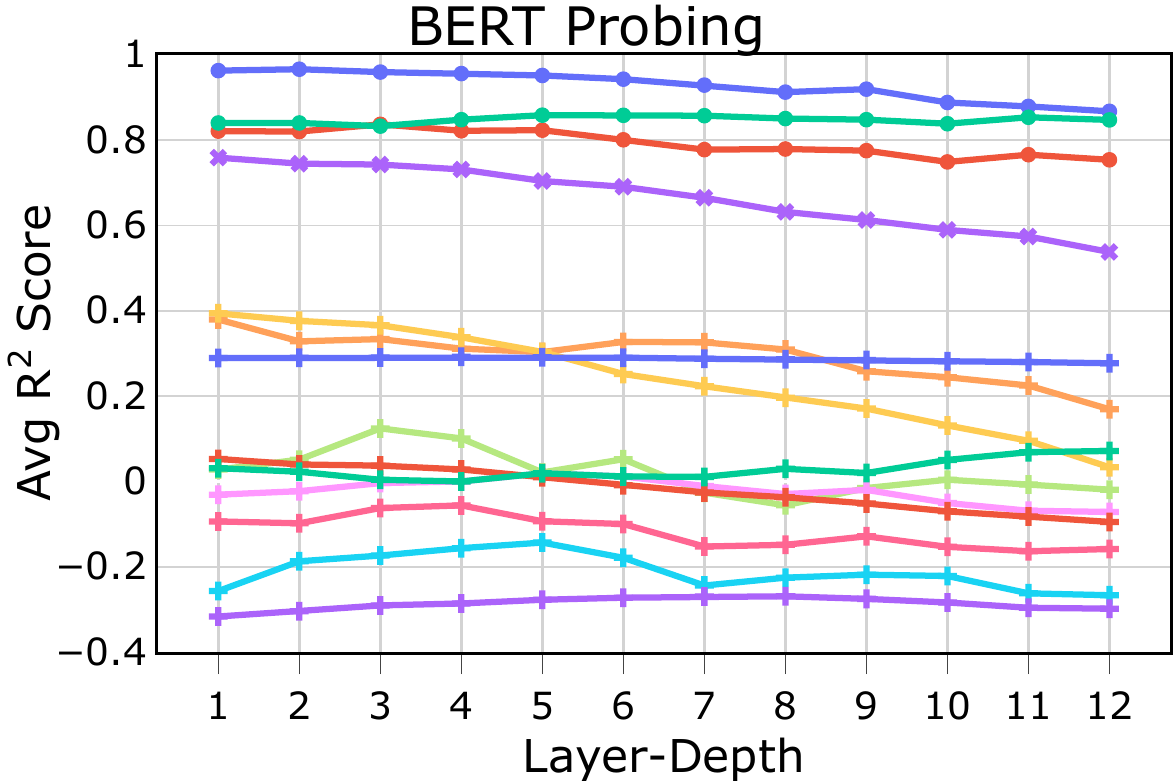}
\includegraphics[width=0.47\linewidth]{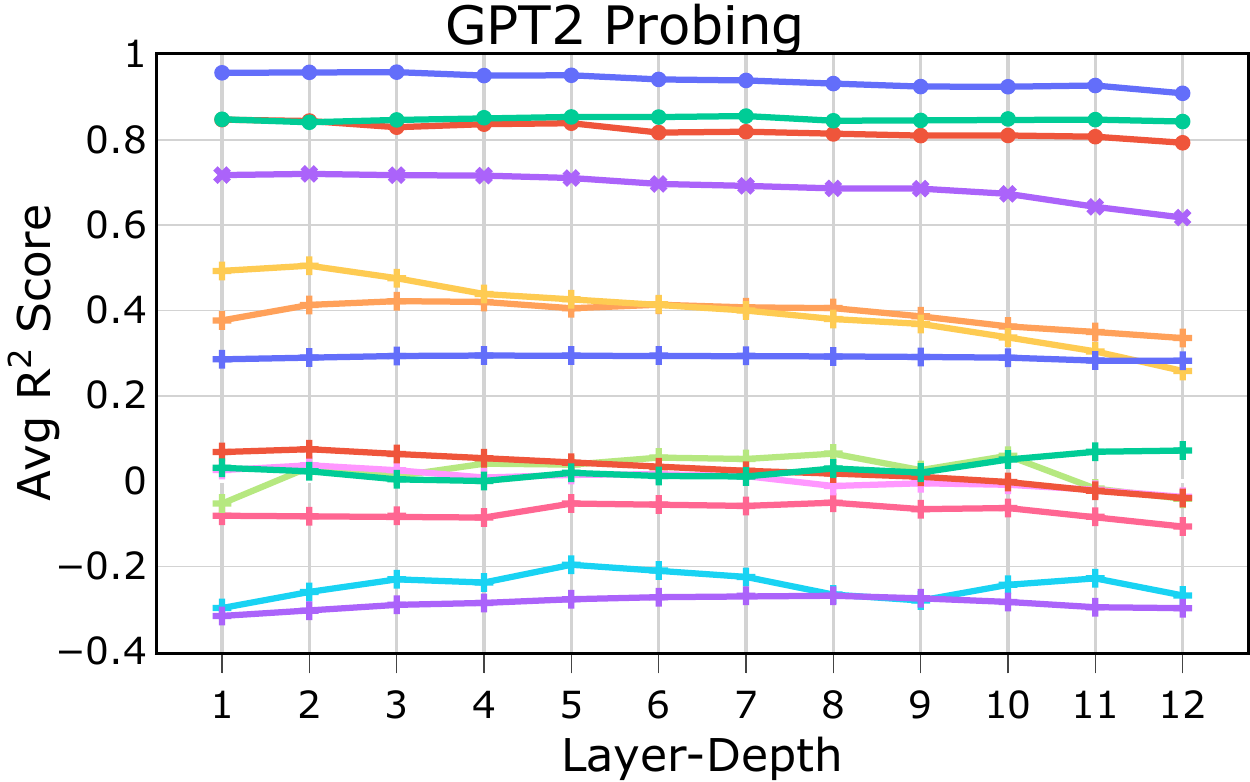}
\includegraphics[width=0.47\linewidth]{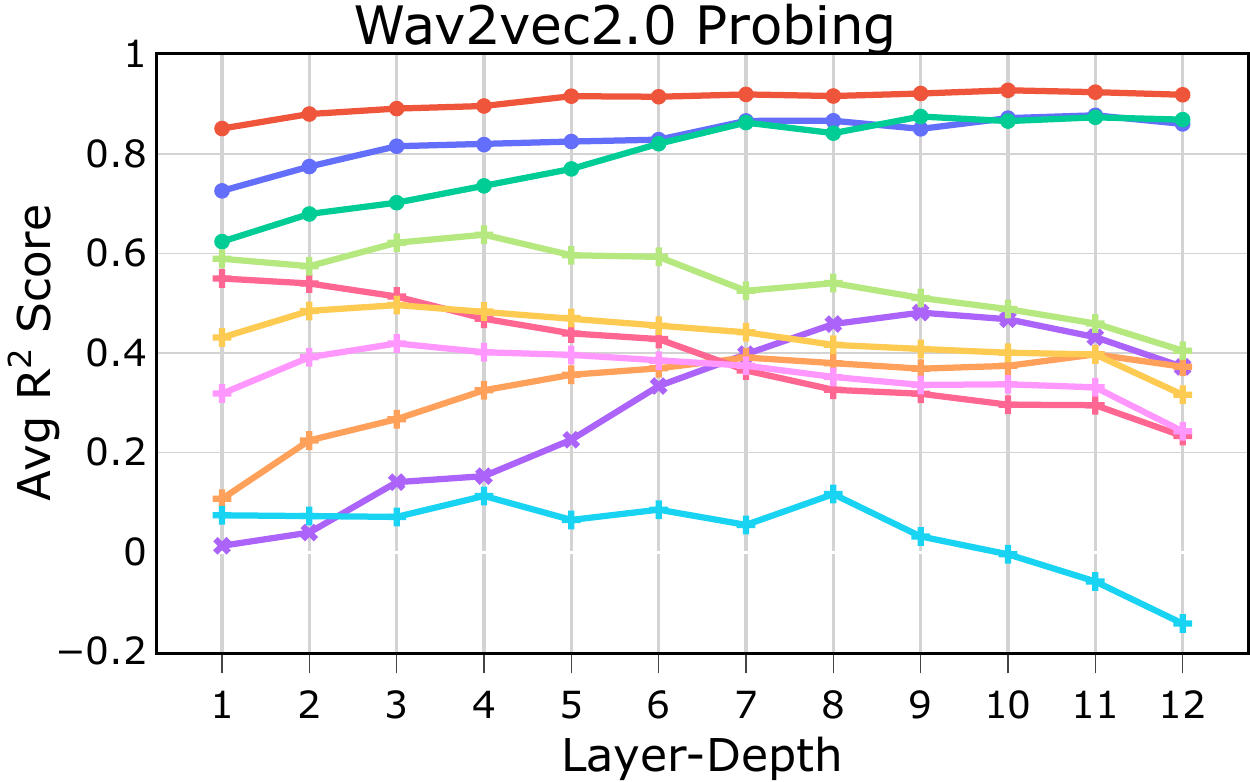}
\includegraphics[width=0.47\linewidth]{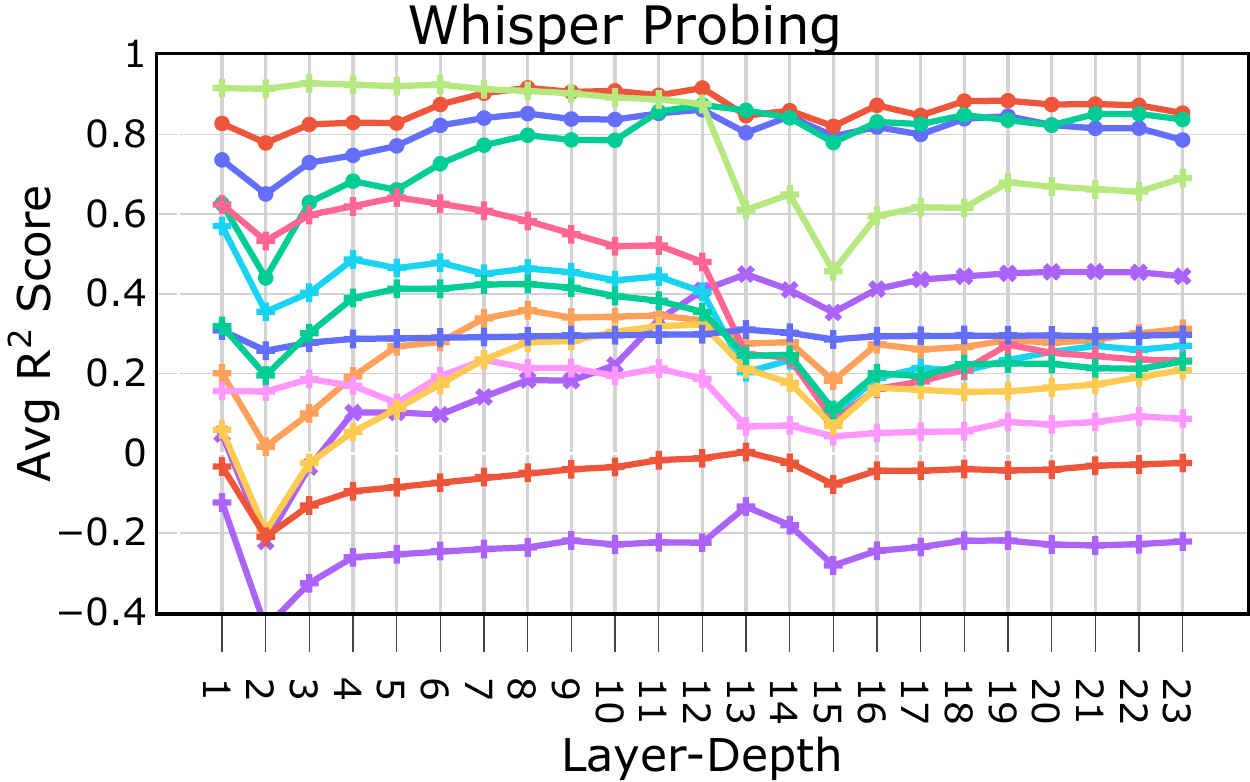}
\caption{Probing the information (basic linguistic and speech features) represented across layers in neural language (BERT and GPT2) and speech-based models (Wav2Vec2.0 and Whisper).}
\label{fig:probing_results_bert_wav2vec}
\end{figure*}

\noindent\textbf{Are there any differences between text- and speech-based models in late language regions?}

\noindent\textbf{Removal of low-level textual features}
In both reading and listening conditions, we make the following observations from  Fig.~\ref{fig:normalized_predictivity_listening_properties} (a) \& (b) and Fig.~\ref{fig:normalized_predictivity_reading_properties} (a) \& (b): (1) Text-based models explain a large amount of variance in late regions, even after removing low-level textual features. (2) In contrast, residual performance of speech-based
models goes down to approximately 10-15\%, after removing number of letters and words. 

\noindent\textbf{Removal of low-level speech features}
In both reading and listening conditions, we make the following observations from Fig.~\ref{fig:normalized_predictivity_listening_properties} (c) \& (d) and Fig.~\ref{fig:normalized_predictivity_reading_properties} (c) \& (d):
(1) Removing DiPhones features from Text-based language models results in major drop (more than 25\%) compared to other low-level speech features.  
(2) Conversely, removal of phonological features results in a larger performance drop (more than 80\% of the original performance) for speech-based language models than text-based models (10\% drop of the original performance).

Overall, the alignment of speech-based models with late language regions is almost entirely due to low-level stimulus features, and not brain-relevant semantics.

\begin{figure*}[!ht] 
\centering
\includegraphics[width=\linewidth]{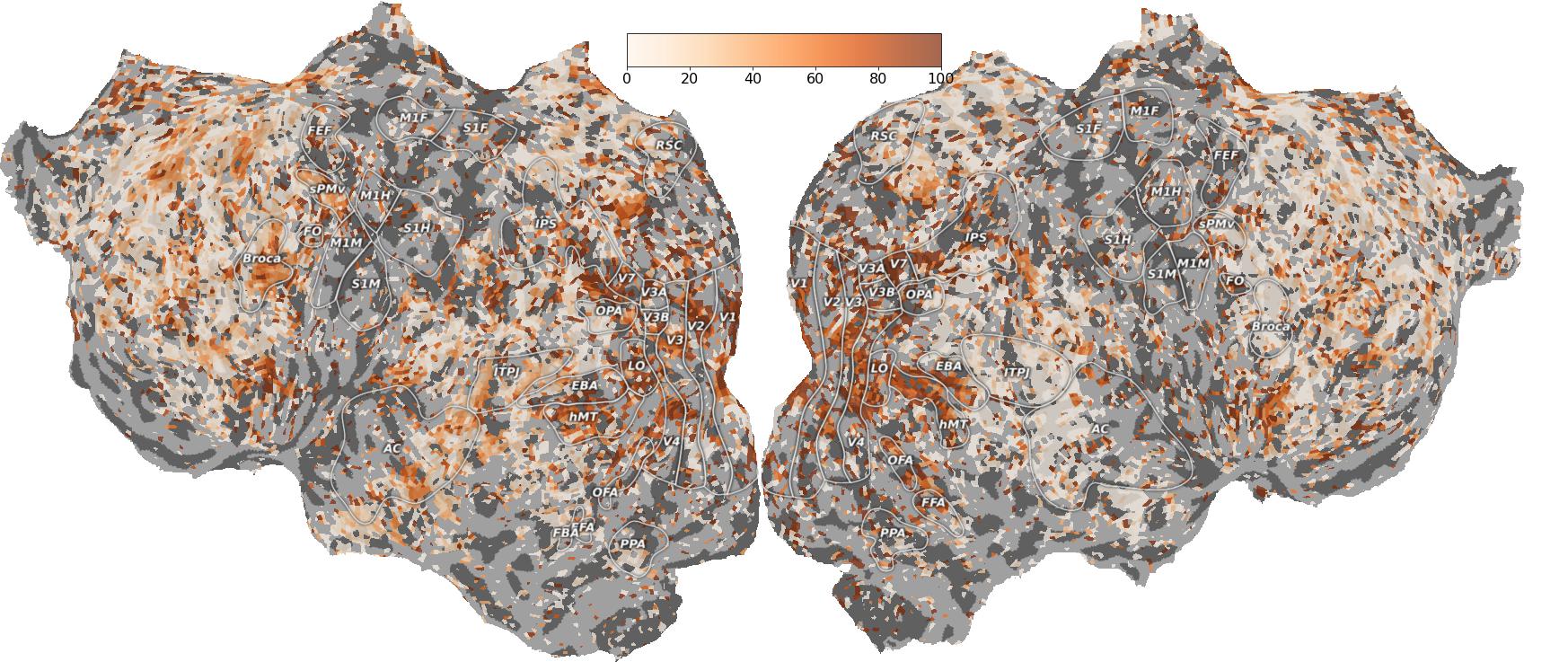}
\caption{Reading: Percentage decrease in alignment for each voxel after removing number of letters feature from BERT representations. Percentage decrease scores for each voxel in one subject (subject-8) are projected onto the subject’s flattened cortical surface. 
Voxels appear White if number of letters feature do not explain any shared information of BERT, and orange if all the information predicted BERT is similar to the information captured using number of letters feature. Voxels appear in light orange indicates that parts of late language explained (0-20\% drop) by number of letters feature and BERT model has more information shared with late language regions
beyond number of letters feature. }
\label{fig:2dcolormap_bert_residual_numletters}
\end{figure*}

\begin{figure*}[!ht] 
\centering
\includegraphics[width=\linewidth]{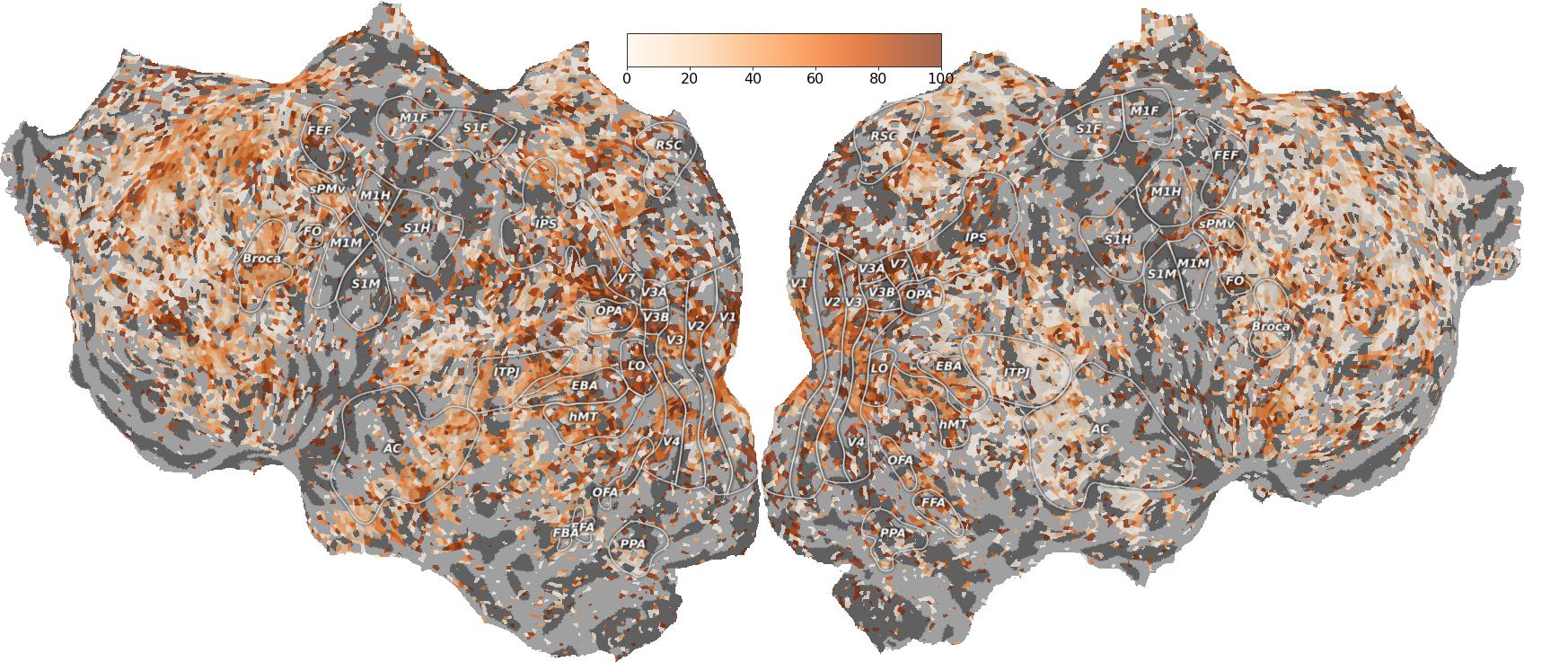}
\caption{Brain Reading: Percentage decrease in brain alignment for each voxel by comparing the results after removing DiPhone features from BERT with the results before using BERT. Percentage decrease scores for each voxel in one subject (subject-8) are projected onto the subject’s flattened cortical surface. 
Voxels appear White if DiPhone features do not explain any shared information of BERT, and orange if all the information predicted BERT is similar to the information captured using DiPhone features. Voxels appear in light orange indicates that parts of late language explained (20-40\% drop) by DiPhone features and BERT model has more information shared with late language regions
beyond DiPhone features. }
\label{fig:2dcolormap_bert_residual_diphone}
\end{figure*}

\section{Layer-wise probing analysis between language models and low-level stimulus features}
To investigate how much of the information in the low-level stimulus features can be captured by text- and speech-based language models, we learn a ridge regression model using model representations as input features to predict the low-level features (textual, speech and visual) as target. Fig.~\ref{fig:probing_results_bert_wav2vec} shows that text-based language model (BERT and GPT-2) can accurately predict low-level textual features in the early layers and have decreasing trend towards later layers. For the low-level speech features, text-based models have zero to negative $R^2$-score values showing that text-based models do not have any speech-level information.

Complementary to text-based language models, speech-based models (Wav2Vec2.0) can accurately predict low-level speech features in the higher layers and have lower $R^2$-score values in the early layers. Conversely, in the Whisper model, which is an encoder-decoder architecture, the encoder layers can accurately predict low-level speech features and decoder layers have higher $R^2$-score values for the low-level textual features.

\begin{figure*}[!ht] 
\centering
\includegraphics[width=0.9\linewidth]{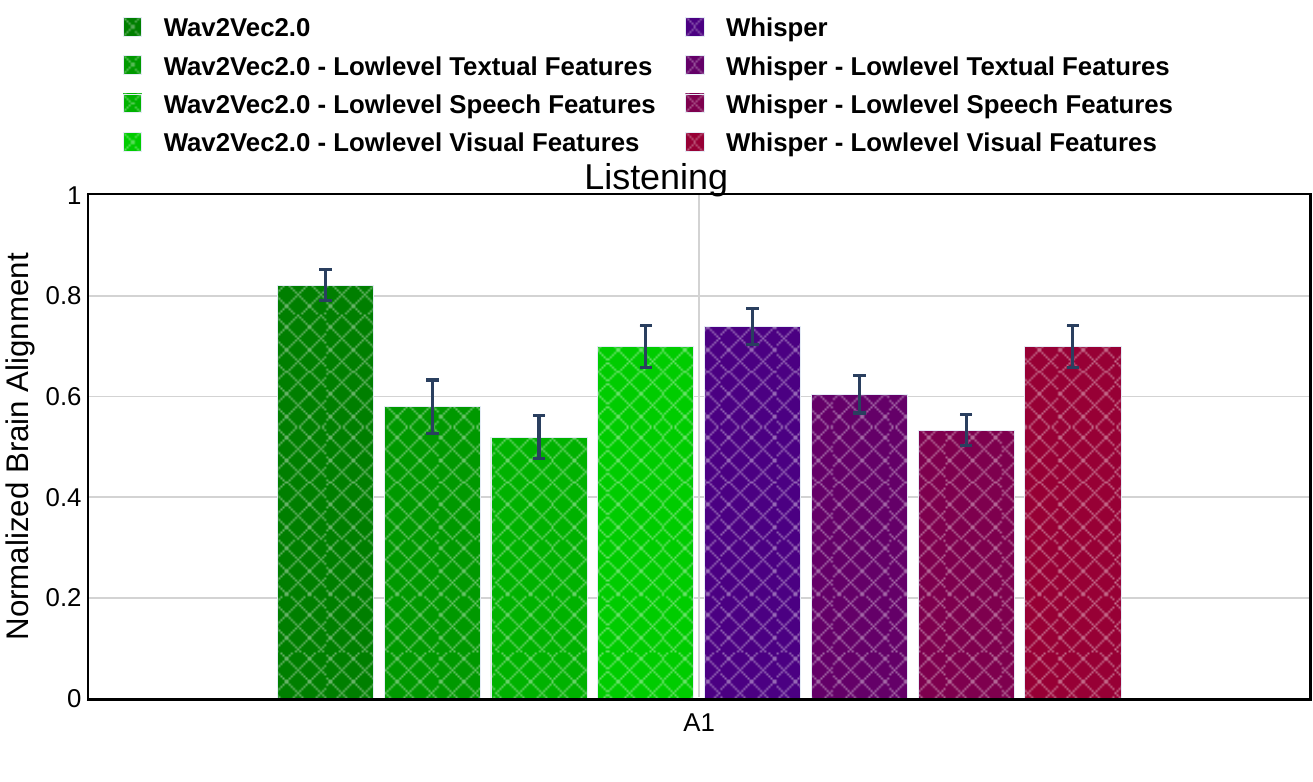}
\includegraphics[width=0.9\linewidth]{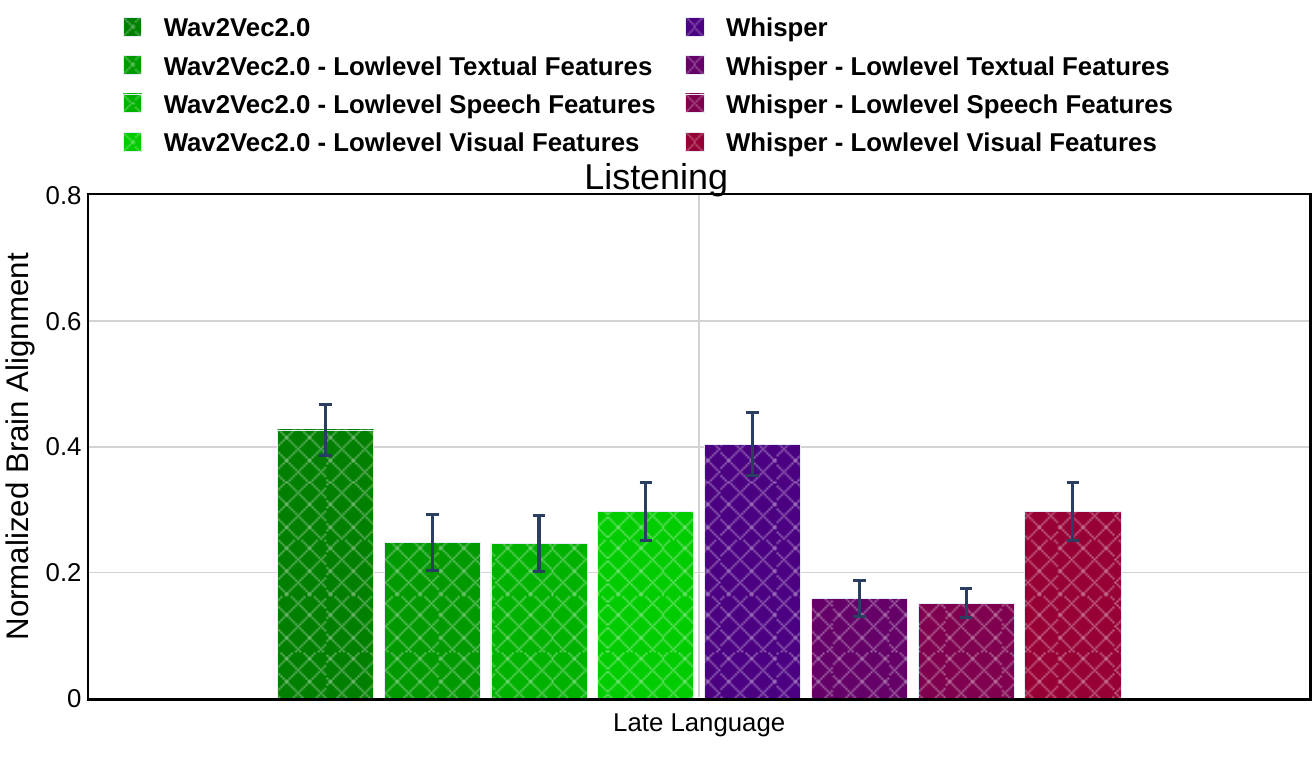}
\caption{Brain Listening in A1 and Late Language regions: Average normalized brain alignment was computed over the average of participants for speech-based models, across layers and across low-level features.}
\label{fig:normalized_predictivity_a1_speechmodels}
\end{figure*}

\section{Feature-level analysis}
As shown in Fig.~\ref{fig:normalized_predictivity_reading_properties}, we observe that "Number of Letters" had the highest impact during reading for the text-based language models.
Fig.  ~\ref{fig:2dcolormap_bert_residual_numletters} displays the percentage decrease in brain alignment for reading (BERT with Number of Letters).
Removing "Number of Letters" leads to a significant drop (80-100\%) in the early visual regions, but only a slight drop (0-20\%) in the late language regions during reading. Fig.~\ref{fig:2dcolormap_bert_residual_diphone} displays the percentage decrease in alignment when the low-level speech feature "DiPhones" is removed from BERT representations during reading.
Since many common short words are composed of diphones ~\citep{gong2023phonemic}, removing this feature from BERT significantly decreases alignment (20-40\%) even in late language regions.


\section{A1 and Late Language regions: Speech-language model alignment during listening}

We now show the results per speech model in the Fig.~\ref{fig:normalized_predictivity_a1_speechmodels}. In the context of brain listening, specifically for the A1 region, we observe that the Wav2Vec2.0 model has better normalized brain alignment than the Whisper model. However, removal of low-level textual and speech features lead to major performance drop in Wav2Vec2.0 than Whisper model. This implies that the maximum explainable variance of Wav2Vec2.0 is due to lower-level features than the Whisper model in A1 region.

Similar to the A1 region, we observed that both Wav2Vec2.0 and Whisper exhibit similar normalized brain alignment in late language regions. Moreover, the removal of low-level textual and speech features results in a significant performance decline in both models.

\section{Layer-wise Normalized Brain Alignment}
\begin{figure*}[!ht] 
\centering
\includegraphics[width=0.65\linewidth]{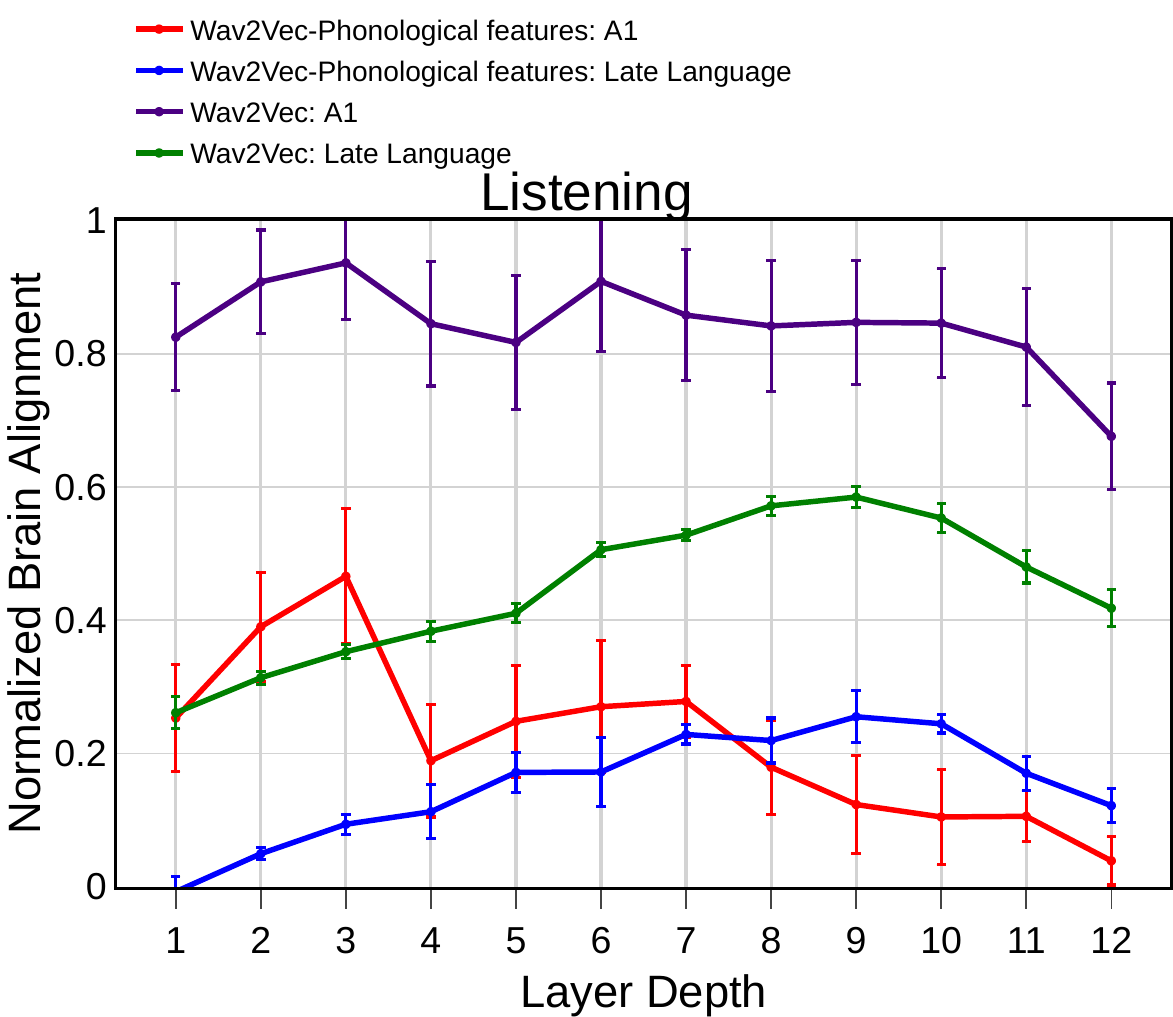}
\caption{Layer-wise average normalized brain alignment was computed over the average of participants for speech-based model: Wav2vec2.0 for an imporant low-level speech property: phonological features.}
\label{fig:wav2vec_layerwise_normalized_predictivity_a1_phonological}
\end{figure*}

We now plot the layer-wise normalized brain alignment for the Wav2Vec2.0 model in brain listening, both before and after removal of one important low-level speech property:  phonological features, as shown in Fig.~\ref{fig:wav2vec_layerwise_normalized_predictivity_a1_phonological}. 
Observation from Fig.~\ref{fig:wav2vec_layerwise_normalized_predictivity_a1_phonological} indicates a consistent drop in performance across layers, after removal of Phonological features, specifically in A1 and Late language regions. 
The key finding here is that our results that low level features impact the ability to predict both A1 and late language regions hold across individual layers.

\section{A1 Region: Low-level stimulus features and brain alignment}

We now plot the average normalized brain alignment for low-level stimulus features (textual, speech and visual) during both reading and listening in the early sensory areas (early visual and A1), as shown in Figure. Additionally, we report the individual low-level stimulus features, such as the number of letters, PowSpec, Phonological features and motion energy features, specifically in early sensory processing regions. It appears that both text-based and speech-based language models meet the baselines in early sensory processing regions, particularly early visual areas in reading and A1 areas during listening. Among low-level stimulus features, motion energy features have better normalized brain alignment during reading in the early visual area and Phonological features have better brain alignment during listening in the A1 region.

\section{Low-level stimulus features: Normalized brain alignment}
\begin{figure*}[!ht] 
\centering
\begin{minipage}{0.9\textwidth}
\includegraphics[width=\linewidth]{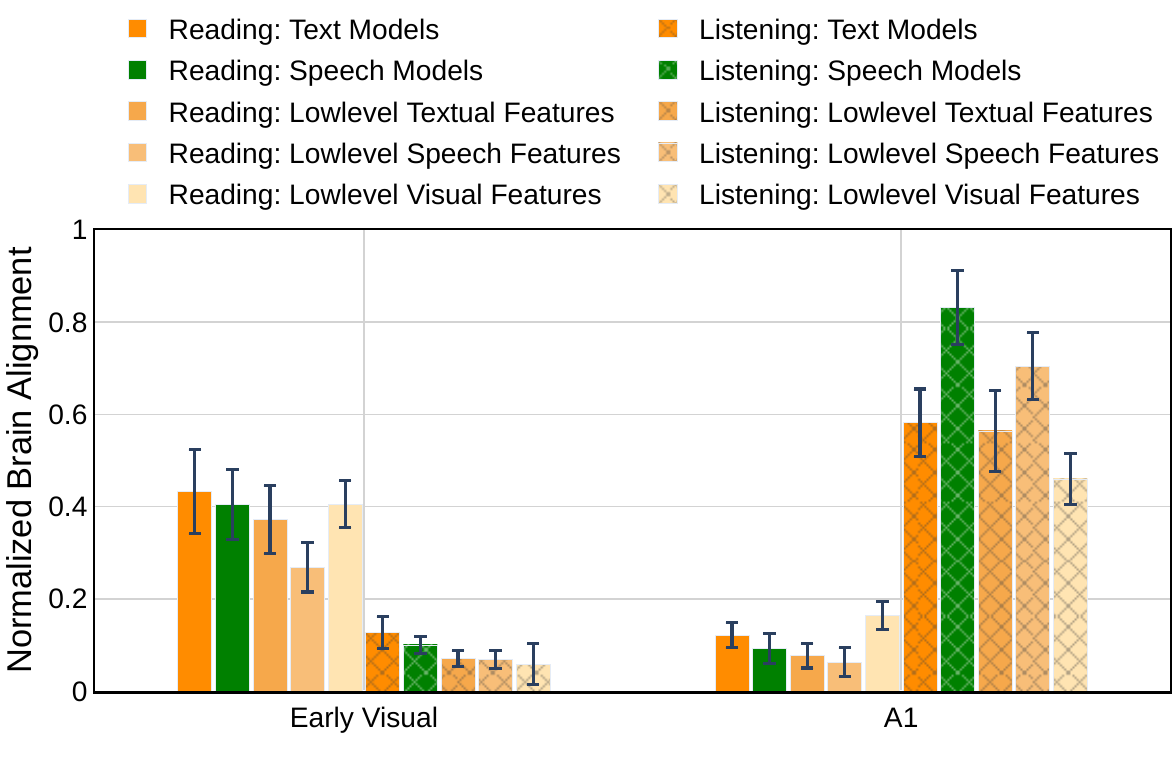}
\\(a) Average of normalized brain alignment of Low-level stimulus features\\
\end{minipage}
\begin{minipage}{0.9\textwidth}
 \includegraphics[width=\linewidth]{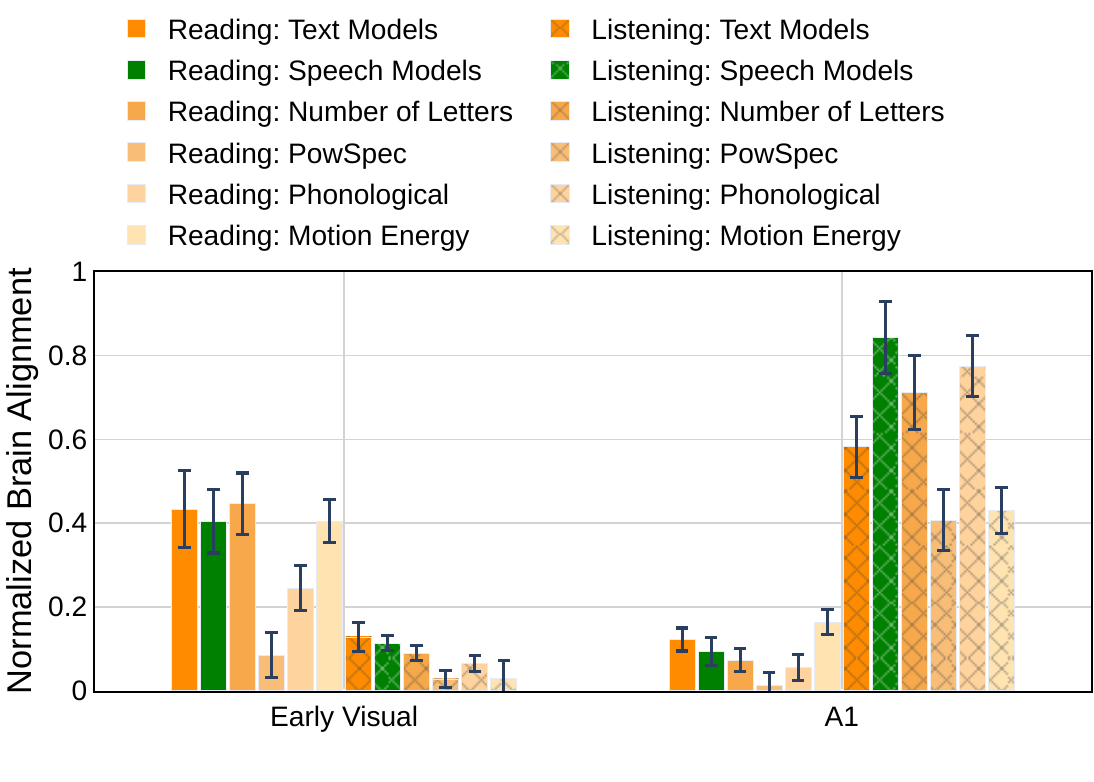}
\\(b) Average of normalized brain alignment of important Low-level stimulus feature\\
 \end{minipage}
\caption{Average normalized brain alignment was computed over the average of participants across (a) text and speech-based models (b) low-level important features}
\label{fig:reading_listening_normalized_predictivity_a1_lowlevl}
\end{figure*}

We now plot the average normalized brain alignment for classical models, such as low-level stimulus features (textual, speech, and visual) during both reading and listening in the early sensory areas (early visual and A1), as shown in Fig.~\ref{fig:reading_listening_normalized_predictivity_a1_lowlevl}. Additionally, we report results for the individual low-level stimulus features as baseline models, including the number of letters, PowSpec, phonological features, and motion energy features, particularly in early sensory processing regions. Both text-based and speech-based language models meet the baselines and show improvement in early sensory processing regions, particularly early visual areas in reading and A1 areas during listening. Among low-level stimulus features, motion energy features have better normalized brain alignment during reading in the early visual area and phonological features have better brain alignment during listening in the A1 region.

Overall, in the context of classic model comparison, language model representations predict better than baseline models and the variance explained by these models is significant.

\begin{figure*}[!ht] 
\centering
\begin{minipage}{0.7\textwidth}
\includegraphics[width=\linewidth]{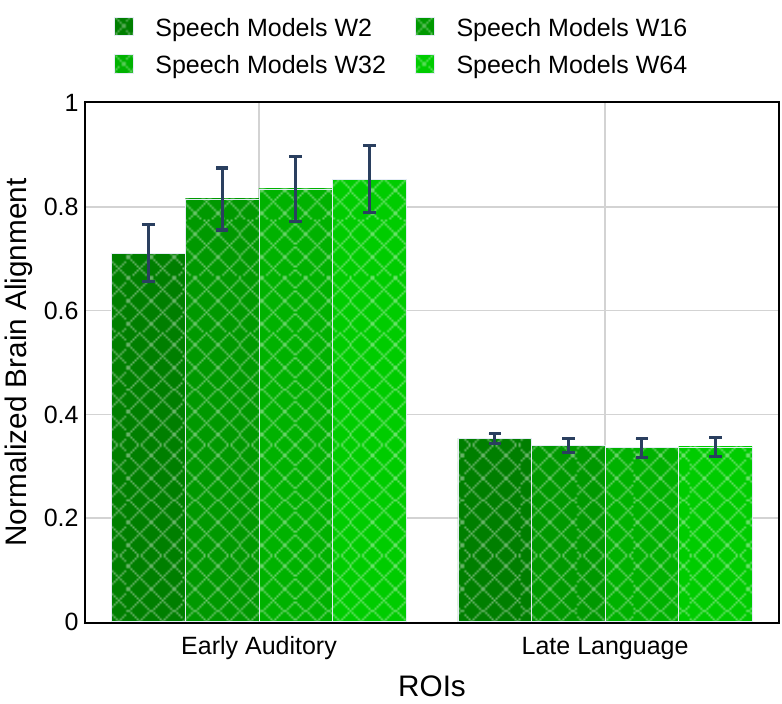}
\\(a) Average of normalized brain alignment was computed over the average of participants across speech-based models for different windows\\
\end{minipage}
\begin{minipage}{0.7\textwidth}
 \includegraphics[width=\linewidth]{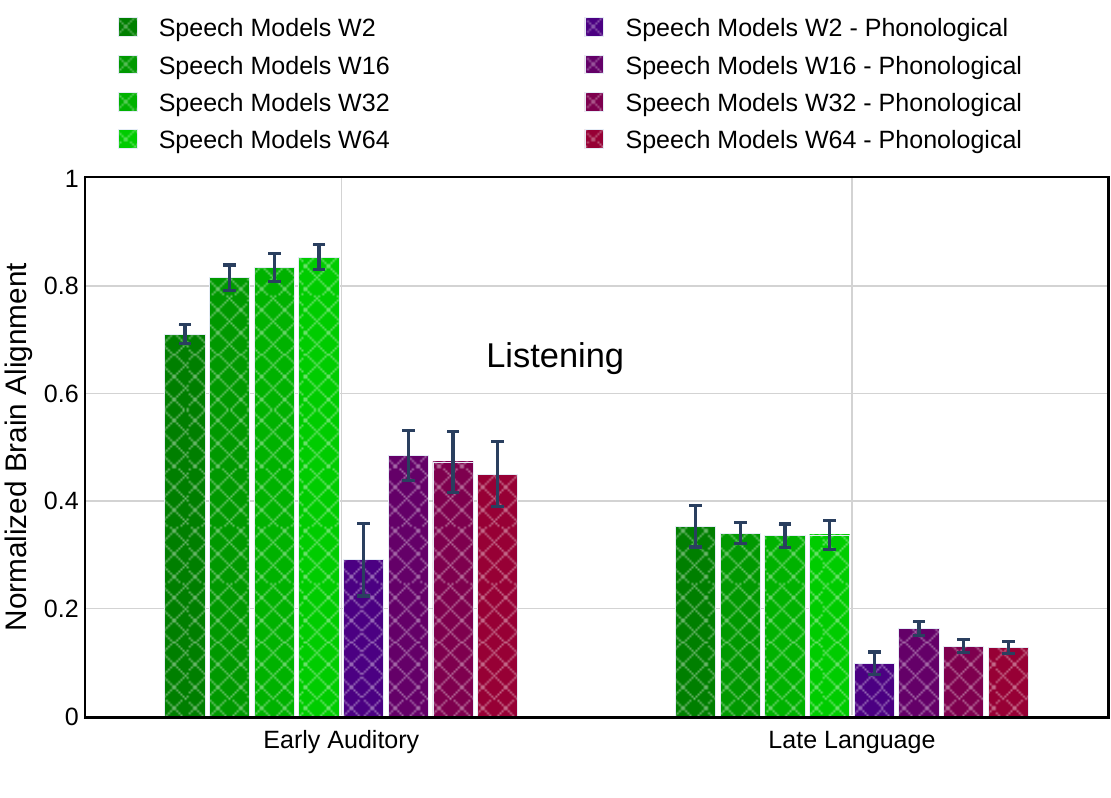}
\\(b) Removal of low-level speech feature: phonological features\\
 \end{minipage}
\caption{Brain Listening: (a) Average normalized brain alignment was computed over the average of participants for speech-based models, (b) Average of normalized brain alignment of speech models for different windows across layers and removal of phonological features }
\label{fig:reading_listening_normalized_predictivity_speech_windows}
\end{figure*}

\begin{table*}[t]
\scriptsize
\begin{center} 

\resizebox{\textwidth}{!}{\begin{tabular}{|c|c|c|c|c|c|c|} 
\hline
\textbf{Model Name} & \textbf{Pretraining}  & \textbf{Type} & \textbf{Input} & \textbf{Layers} & \textbf{Training data size} & \textbf{Loss type}
 \\
\hline
BERT-base-uncased  & Text&  Encoder (Bidirectional) & Words& 12 &3.3B words & Masked language model  (MLM) \\
GPT2-Small &Text &   Decoder (Unidirectional) & Words & 12 &8M web pages &Causal language model (CLM) \\
FLAN-T5-base& Text&Encoder-Decoder & Words & 24 & 750GB corpus of tex&MLM, CLM\\
Wav2Vec2.0-base&Speech& Encoder & Waveform & 12 & 250K hours of raw speech  & Masked contrastive loss\\
Whisper-small&Speech& Encoder-Decoder & Log Mel spectrogram & 24 & 680K hours of speech (raw speech+speech tasks) & Masked dynamic loss\\
\hline
\end{tabular} }
\vspace{-0.2cm}
\caption{Pretrained Transformer-based language models} 
\label{neural_models_details} 
\end{center} 
\end{table*}

\section{Impact of the phonological feature.}
\label{phonological_features_impact}

\begin{figure*}[!ht]
    \centering
    \includegraphics[width=\linewidth]{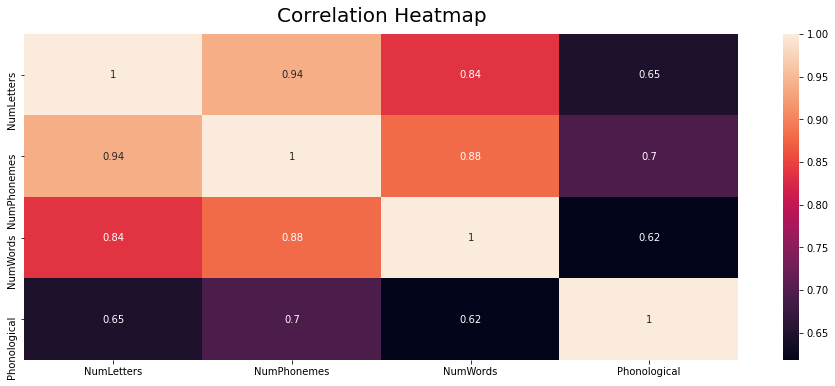}
    \caption{Correlation matrix illustrating the relationships between low-level features: NumLetters, NumPhonemes, NumWords, and Phonological.}
    \label{fig:correlation_phonological}
\end{figure*}

\section{Functionality of
brain ROIs}
We use a multi-modal parcellation of the human cerebral cortex (Glasser Atlas; 180 regions of interest (ROIs) in each hemisphere)~\citep{glasser2016multi}. This includes two early sensory processing regions and seven language-relevant ROIs in the human brain with the following subdivisions: (1) early visual (V1, V2); (2) early auditory region (A1, PBelt, MBelt, LBelt, RI, A4); and (3) late language regions, encompassing broader language regions: angular gyrus (AG: PFm, PGs, PGi, TPOJ2, TPOJ3), lateral temporal cortex (LTC: STSda, STSva, STGa, TE1a, TE2a, TGv, TGd, A5, STSdp, STSvp, PSL, STV, TPOJ1), inferior frontal gyrus (IFG: 44, 45, IFJa, IFSp) and middle frontal gyrus (MFG: 55b)~\citep{baker2018connectomic,milton2021parcellation,desai2022proper}.
In Table~\ref{rois_description}, we report the comprehensive description of early sensory brain regions and language-relevant ROIs in the human brain with the following subdivisions: (1)
early visual; (2) visual word form area; (3) early auditory region; and (4) late language regions.

\section{Speech-based Language Models}
Speech-based models are often referred to as language models, because the general definition of a “language model” involves predicting the next token or sequence of tokens, given context. Even in the paper which introduces the Wav2vec2.0 model we use~\citep{baevski2020wav2vec}, the authors state that the model is "pretrained by masking specific time steps in the latent feature space, similar to the masked language modeling approach in BERT". Although the Wav2vec2.0 model does not generate text in the way traditional text-based language models do, it plays a crucial role in processing and understanding spoken language, making it a type of speech language model. Additionally, in another work that introduces another popular speech model HuBERT, the authors explicitly state that the "HuBERT model is compelled to learn both acoustic and language models from continuous inputs"~\citep{hsu2021hubert}. 
Whether these speech models capture the same meaning of language as traditional text-based language models is not known. Cognitive neuroscientists are starting to study the overall alignment of such speech models with the human brain, across the whole cortex and not just in early auditory areas, so our work is extremely timely in showing that the observed alignment in non-sensory regions is largely due to low-level features.

The details of three text-based and two speech-based language models are reported in Table~\ref{neural_models_details}.

\setlength{\tabcolsep}{3pt}
\begin{table*}[!h]
\centering
\begin{tabular}{|p{1in}|p{4in}|}
\hline
Early visual & The early visual region is the earliest cortical region for visual processing. It processes basic visual features, such as edges, orientations, and spatial frequencies. Lesions in V1 can lead to blindness in the corresponding visual field. V2 processes more complex patterns than V1.\\
\hline
Early auditory & The early auditory region is the earliest cortical region for speech processing. This region is specialized for processing elementary speech sounds, as well as other temporally complex acoustical signals, such as music.\\
\hline
Late Language & Late language regions contribute to various linguistic processes. Regions 44 and 45 (Broca's region) are vital for speech production and grammar comprehension~\citep{friederici2011brain}. The IFJ, PG, and TPOJ clusters are involved in semantic processing, syntactic interpretation, and discourse comprehension~\citep{deniz2019representation,toneva2022combining}. STGa and STS play roles in phonological processing and auditory-linguistic integration~\citep{vaidya2022self,millet2022toward,gong2023phonemic}. TA2 is implicated in auditory processing, especially in the context of language. \\
\hline
\end{tabular}
\caption{Detailed functional description of various brain regions.}
\label{rois_description}
\end{table*}

\end{document}